\documentclass[10pt,twocolumn,letterpaper]{article}

\usepackage{cvpr}
\usepackage{times}
\usepackage{epsfig}
\usepackage{graphicx}
\usepackage{amsmath}
\usepackage{amssymb}

\usepackage{xcolor}

\usepackage{authblk}

\usepackage{multirow}

\usepackage{makecell}	

\usepackage[export]{adjustbox}

\usepackage{diagbox}
\usepackage{numprint}

\usepackage{verbatim}
\usepackage{pgfplots} 	

\usepackage{tikz}
\usetikzlibrary{shapes,arrows,positioning,calc,patterns}

\usepackage{mathtools}
\usepackage{caption} 

\usepackage[labelformat=simple]{subcaption}

\usepackage{bm}	
\renewcommand{\vec}[1]{\bm{\mathbf{#1}}}

\usepackage[breaklinks=true,bookmarks=false]{hyperref}

\cvprfinalcopy 

\def\cvprPaperID{****} 

\begin{document}

\graphicspath{{./images/dataset/}{./images/overview/}{./images/quantevaluation/}{./images/realimages/}}

\title{3D Trajectory Reconstruction of Dynamic Objects Using Planarity Constraints}

\renewcommand{\Authands}{ and } 

\author[1]{Sebastian Bullinger}
\author[1]{Christoph Bodensteiner}
\author[1]{Michael Arens}
\author[2]{Rainer Stiefelhagen}
\affil[1]{Fraunhofer IOSB}
\affil[2]{Karslruhe Institute of Technology}
\affil[ ]{\tt\small \textit {\{sebastian.bullinger,christoph.bodensteiner,michael.arens\}@iosb.fraunhofer.de}}
\affil[ ]{\tt\small \textit rainer.stiefelhagen@kit.edu}

\maketitle

\begin{abstract}
	We present a method to reconstruct the three-dimensional trajectory of a moving instance of a known object category in monocular video data. We track the two-dimensional shape of objects on pixel level exploiting instance-aware semantic segmentation techniques and optical flow cues. We apply Structure from Motion techniques to object and background images to determine for each frame camera poses relative to object instances and background structures. By combining object and background camera pose information, we restrict the object trajectory to a one-parameter family of possible solutions. We compute a ground representation by fusing background structures and corresponding semantic segmentations. This allows us to determine an object trajectory consistent to image observations and reconstructed environment model. Our method is robust to occlusion and handles temporarily stationary objects. We show qualitative results using drone imagery. Due to the lack of suitable benchmark datasets we present a new dataset to evaluate the quality of reconstructed three-dimensional object trajectories. The video sequences contain vehicles in urban areas and are rendered using the path-tracing render engine Cycles to achieve realistic results. We perform a quantitative evaluation of the presented approach using this dataset. Our algorithm achieves an average reconstruction-to-ground-truth distance of 0.31 meter. The dataset will be publicly available on our website\footnote{\label{project_page}Project page: URL}.
\end{abstract}

\newcommand\firstcol{0}
\newcommand\firstrow{0}

\newcommand\secondcol{6}
\newcommand\secondrow{-3}

\newcommand\thirdcol{12}
\newcommand\thirdrow{-6}

\newcommand\fourthrow{-9}

\newcommand\includegraficsFirstRowDistanceRatio{0.5}
\newcommand\overviewVerticalShiftFirstRow{-0.25}
\newcommand\overviewVerticalShiftSecondToThirdRow{0.25}
\newcommand\overviewVerticalShiftThirdRow{-0.25}

\newlength{\includegraficsWidthOverview}
\setlength{\includegraficsWidthOverview}{1in}

\newlength{\includegraficsHeightOverview}
\setlength{\includegraficsHeightOverview}{0.5in}

\newlength{\subfigureWidthTwoColumns}
\setlength{\subfigureWidthTwoColumns}{2.4in}

\newlength{\subfigureWidthThreeColumns}
\setlength{\subfigureWidthThreeColumns}{1.6in}

\newlength{\subfigureWidthFourColumns}
\setlength{\subfigureWidthFourColumns}{1.65in}

\section{Introduction}
\label{section:introduction}

\subsection{Trajectory Reconstruction}
\label{subsection:trajectory_reconstruction}

The reconstruction of three-dimensional object motion trajectories is important for autonomous systems and augmented reality applications. There are different platforms like drones or wearable systems where one wants to achieve this task with a minimal number of devices in order to reduce weight or lower production costs. We propose an approach to reconstruct three-dimensional object motion trajectories using a single camera as sensor. \\
The reconstruction of object motion trajectories in monocular video data captured by moving cameras is a challenging task, since in general it cannot be solely solved exploiting image observations. Each observed object motion trajectory is scale ambiguous. Additional constraints are required to identify a motion trajectory consistent to background structures. 
\cite{Song2016,Lee2015,Chhaya2016} assume that the camera is mounted on a driving vehicle, i.e. the camera has specific height and a known pose. \cite{Ozden2004a,Yuan2006,Park2015} solve the scale ambiguity by making assumptions about object and camera motion trajectories. We follow Ozden's principle of non-accidental motion trajectories \cite{Ozden2004a} and introduce a new object motion constraint exploiting semantic segmentation and terrain geometry to compute consistent object motion trajectories. \\
In many scenarios objects cover only a minority of pixels in video frames. This increases the difficulty of reconstructing object motion trajectories using image data. In such cases current state-of-the-art Structure from Motion (SfM) approaches  treat moving object observations most likely as outliers and reconstruct background structures instead. Previous works, e.g. \cite{Kundu2011, Lebeda2014}, tackle this problem by considering multiple video frames to determine moving parts in the video. They apply motion segmentation or keypoint tracking to detect moving objects. These kind of approaches are vulnerable to occlusion and require objects to move in order to separate them from background structures. \\
Our method exploits recent results in instance-aware semantic segmentation and rigid Structure from Motion techniques. Thus, our approach extends naturally to stationary objects. In addition, we do not exploit specific camera pose constraints like a fixed camera-ground-angle or a fixed camera-ground-distance. We evaluate the presented object motion trajectory reconstruction algorithm in UAV scenarios, where such constraints are not valid.

\subsection{Related Work}
\label{subsection:related_work}

Semantic segmentation or scene parsing is the task of providing semantic information at pixel-level. Early semantic segmentation approaches using ConvNets, e.g.  Farabet et al. \cite{farabet2013}, exploit patchwise training. Long et al. \cite{Shelhamer2017} applied Fully Convolutinal Networks for semantic segmentation, which are trained end-to-end. Recently, \cite{Dai2016,Li2016,HeGDG17} proposed  instance-aware semantic segmentation approaches. \\
The field of Structure from Motion (SfM) can be divided into iterative and global approaches. Iterative or sequential SfM methods \cite{Snavely2006, Wu2011, Moulon2013, Sweeney2014, Schoenberger2016sfm} are more likely to find reasonable solutions than global SfM approaches \cite{Moulon2013, Sweeney2014}. However, the latter are less prone to drift. \\
The determination of the correct scale ratio between object and background reconstruction requires additional constraints. Ozden et al. \cite{Ozden2004a} exploit the non-accidentalness principle in the context of independently moving objects. Yuan et al. \cite{Yuan2006} propose to reconstruct the 3D object trajectory by assuming that the object motion is perpendicular to the normal vector of the ground plane. 
Kundu et al. \cite{Kundu2011} exploit motion segmentation with multibody VSLAM to reconstruct the trajectory of moving cars. They use an instantaneous constant velocity model in combination with Bearing only Tracking to estimate consistent object scales.
Park et al. propose an approach in \cite{Park2015} to reconstruct the trajectory of a single 3D point tracked over time by approximating the motion using a linear combination of trajectory basis vectors. Previous works, like \cite{Ozden2004a,Yuan2006,Kundu2011,Park2015} show only qualitative results.





\subsection{Contribution}

The core contributions of this work are as follows.~(1) We present a new framework to reconstruct the three-dimensional trajectory of moving instances of known object categories in monocular video data leveraging sate-of-the-art semantic segmentation and structure from motion approaches.~(2) We propose a novel method to compute object motion trajectories consistent to image observations and background structures.~(3) In contrast to previous work, we quantitatively evaluate the reconstructed object motion trajectories.~(4) We created a new object motion trajectory benchmark dataset due to the lack of publicly available video data of moving objects with suitable ground truth data. The dataset consists of photo-realistic rendered videos of urban environments. It includes animated vehicles as well as set of predefined camera and object motion trajectories. 3D vehicle and environmental models used for rendering serve as ground truth.~(5) We will publish the dataset and evaluation scripts to foster future object motion reconstruction related research. 

\subsection{Paper Overview}

The paper is organized as follows. Section \ref{section:methods} describes the structure and the components of the proposed pipeline. In section \ref{subsection:trajectory_representation} we derive an expression for a one-parameter family of possible object motion trajectories combining object and background reconstruction results. Section \ref{subsection:ground_computation} describes a method to approximate the ground locally. In section \ref{subsection:scale_estimation_using_constant_distance} we describe a method to compute consistent object motion trajectories. In section \ref{section:experiments_and_evaluation} we provide an qualitative and quantitative evaluation of the presented algorithms using drone imagery and rendered video data. Section \ref{section:conclusions} concludes the paper. 

\section{Object Motion Trajectory Reconstruction}
\label{section:methods}

\begin{figure*}[!tb]
	\centering
	
	\newcommand{\figureTextSize}{\footnotesize}
	

	\tikzstyle{blockRounded} = [rectangle, draw, text width=6em, text centered, rounded corners, font=\figureTextSize]
	\tikzstyle{blockCorner} = [rectangle, draw, text width=6em, text centered, font=\figureTextSize]
	\tikzstyle{line} = [draw, -latex', font=\figureTextSize, text centered, text width=18mm]
	
	\tikzstyle{imageStyle} = [
							text width = \includegraficsWidthOverview,
							text height = \includegraficsHeightOverview]
	\tikzstyle{imageDescriptionStyle} = [
										fill=white, 
										opacity=0.5, 
										text opacity=1, 
										font=\figureTextSize,
										text width = \includegraficsWidthOverview]
	\tikzstyle{plainDescriptionStyle} = [
		rectangle,
		draw,
		fill=white, 
		opacity=1, 
		text opacity=1, 
		font=\figureTextSize,
		text width = \includegraficsWidthOverview]
	
	
	  \begin{tikzpicture}[auto]


		\node (inputFrames) at (\firstcol,\firstrow) {};
	  	\node [blockRounded] (semanticSegmentation)  at (\secondcol,\firstrow)  {Semantic Segmentation and Object Tracking};
	  	\node (dummySegmentationLevelNodeRight) at (\thirdcol,\firstrow)  {};
	
		\node [blockRounded] (objectSfM)  at (\firstcol,\secondrow) {SfM};
	  	\node [blockRounded] (backgroundSfM)  at (\secondcol,\secondrow)  {SfM};
	 	\node (dummySfMLevelNodeRight) at (\thirdcol,\secondrow)  {};
	  
  		\node [blockRounded] (scaleEstimation) at (\secondcol,-5) {Scale Estimation and Trajectory Computation};
	  
		\node [blockRounded] (trajectoryFamilyComputation) at (\firstcol,\thirdrow) {Trajectory Family Computation};
	  	\coordinate (ConsistentObjectTrajectoryCoordinate) at ($(\secondcol,\thirdrow)+(0,-0.25)$);
	  	\coordinate (finalObjectTrajectoryCoordinate) at ($(\secondcol,\thirdrow)+(0,-1)$);
	  	\node [blockRounded] (groundComputation) at (\thirdcol,\thirdrow) {Ground Computation};

	  \begin{scope}[every path/.style=line]
		\draw (inputFrames) -- (semanticSegmentation);
		\node[imageStyle] (framesToTracking) at ($(inputFrames)!0.5!(semanticSegmentation)$) { 
			\includegraphics[ frame, width = \includegraficsWidthOverview]{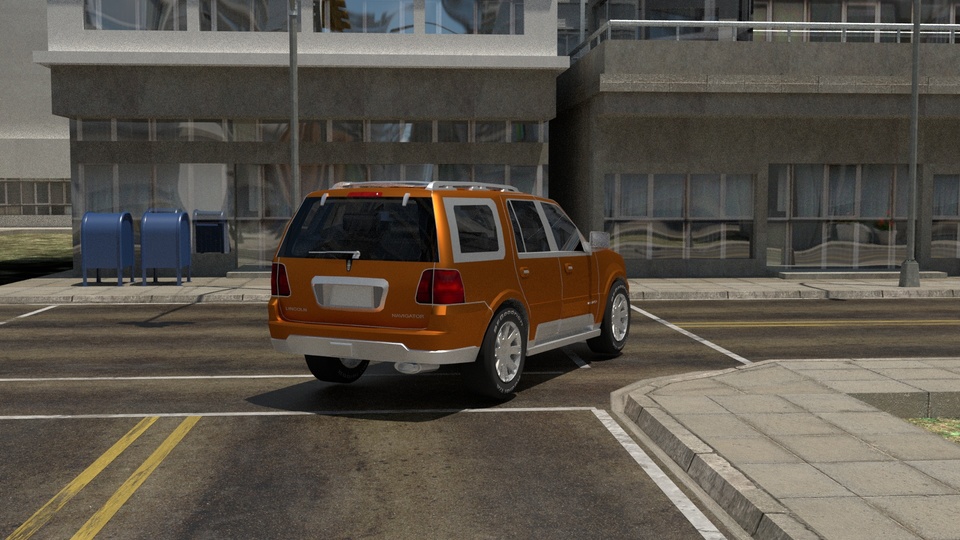}};
		\node[imageDescriptionStyle] at (framesToTracking.center) {Input Frames};

		\draw (semanticSegmentation) -- 
		(objectSfM);
		\node[imageStyle] (sSegmentationToObject) at ($(semanticSegmentation)!\includegraficsFirstRowDistanceRatio!(objectSfM)+(0,\overviewVerticalShiftFirstRow)$) { 
			\includegraphics[ frame, width = \includegraficsWidthOverview]{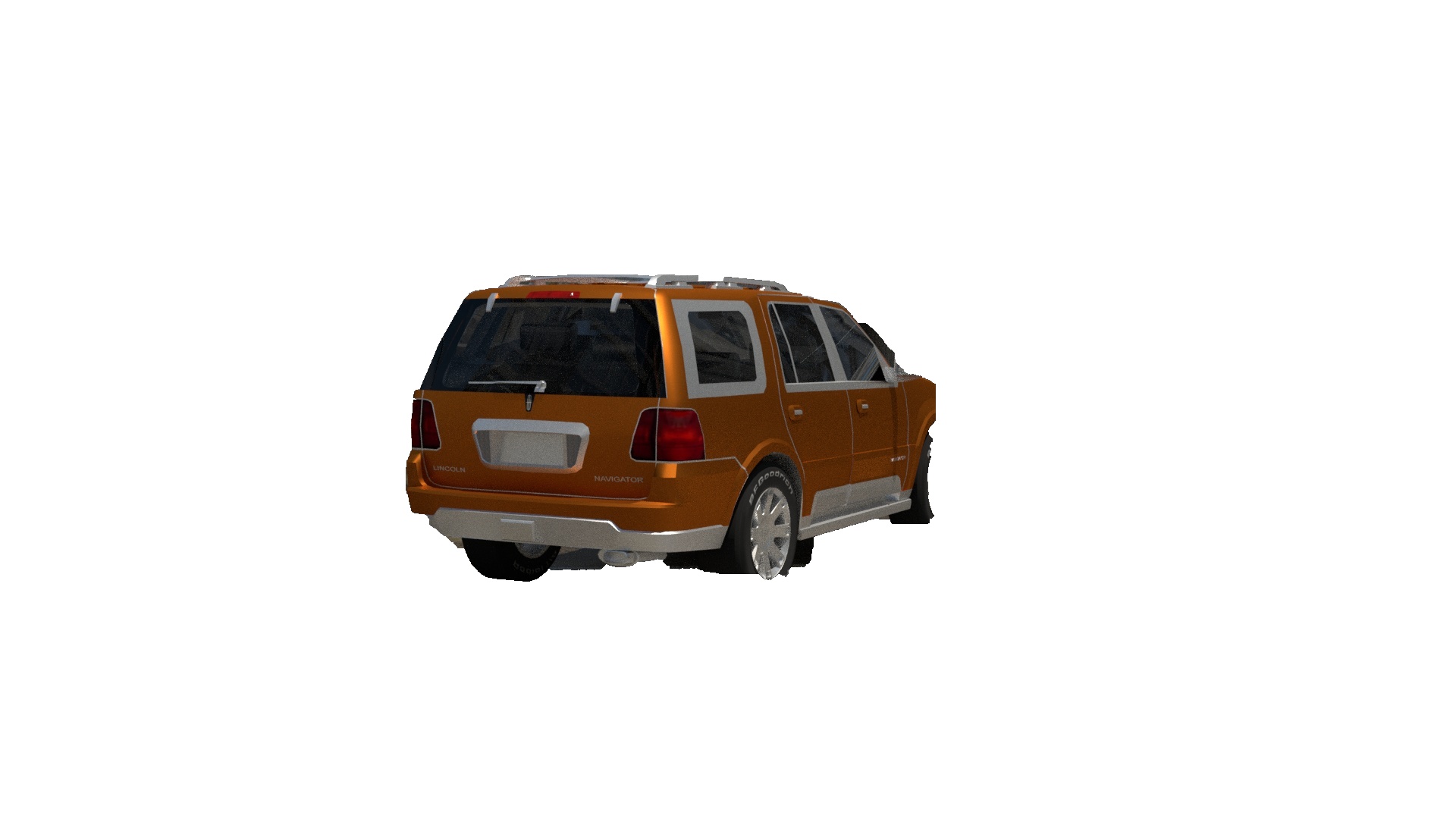}};
		\node[imageDescriptionStyle] at (sSegmentationToObject.center) {Object Segmentations};
		
		\draw (semanticSegmentation) -- (backgroundSfM);
		\node[imageStyle] (sSegmentationToBackgroundImage) at ($(semanticSegmentation)!\includegraficsFirstRowDistanceRatio!(backgroundSfM)+(0,\overviewVerticalShiftFirstRow)$){ 
			\includegraphics[ frame, width = \includegraficsWidthOverview]{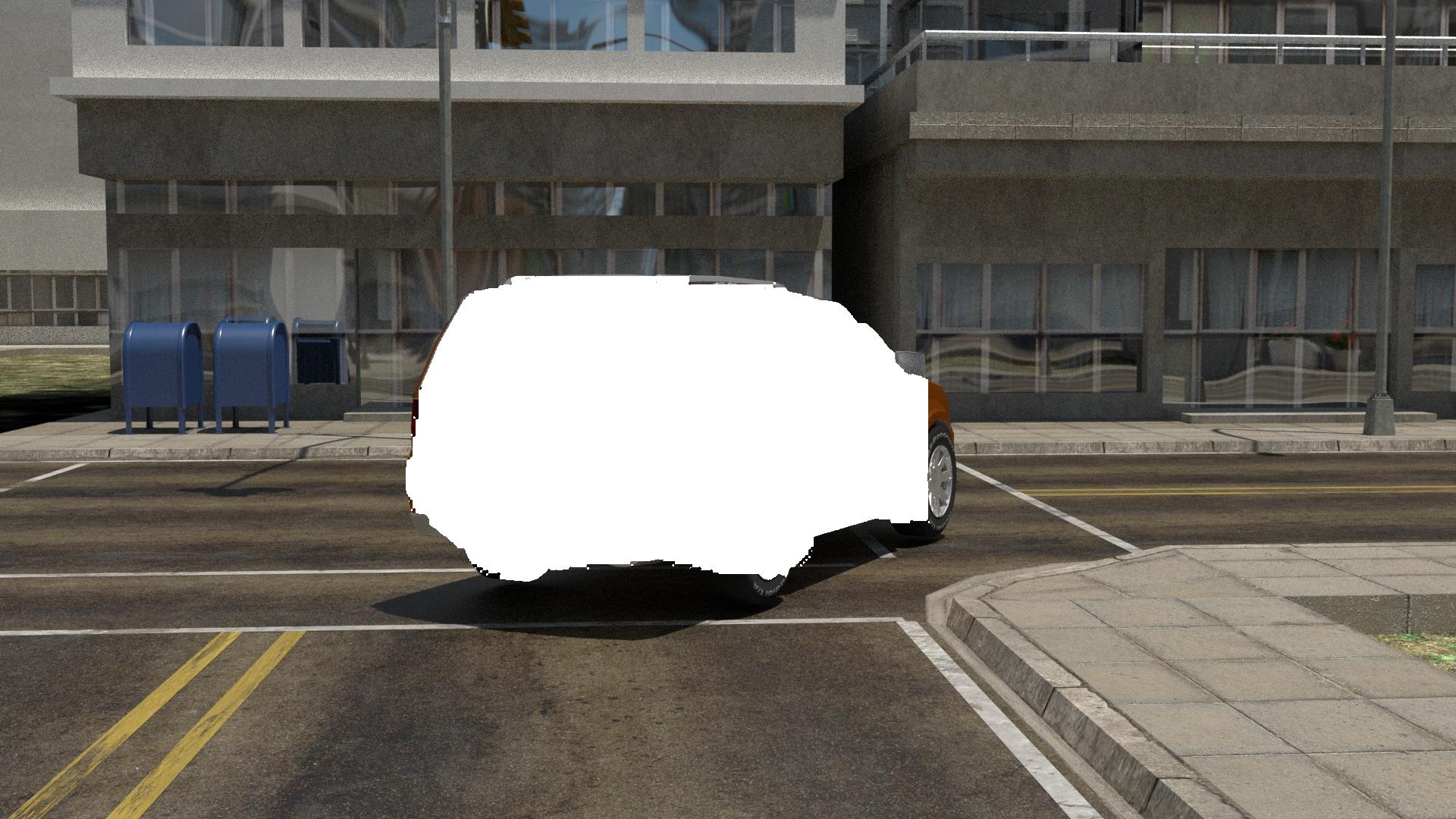}};
		\node[imageDescriptionStyle] (sSegmentationToBackgroundDescription) at (sSegmentationToBackgroundImage.center) {Background Segmentations};
		
		\draw (semanticSegmentation) -- (dummySfMLevelNodeRight); 
		\node[imageStyle] (sSegmentationToGroundImage) at ($(semanticSegmentation)!\includegraficsFirstRowDistanceRatio!(dummySfMLevelNodeRight)+(0,\overviewVerticalShiftFirstRow)$) { 
			\includegraphics[frame, width = \includegraficsWidthOverview]{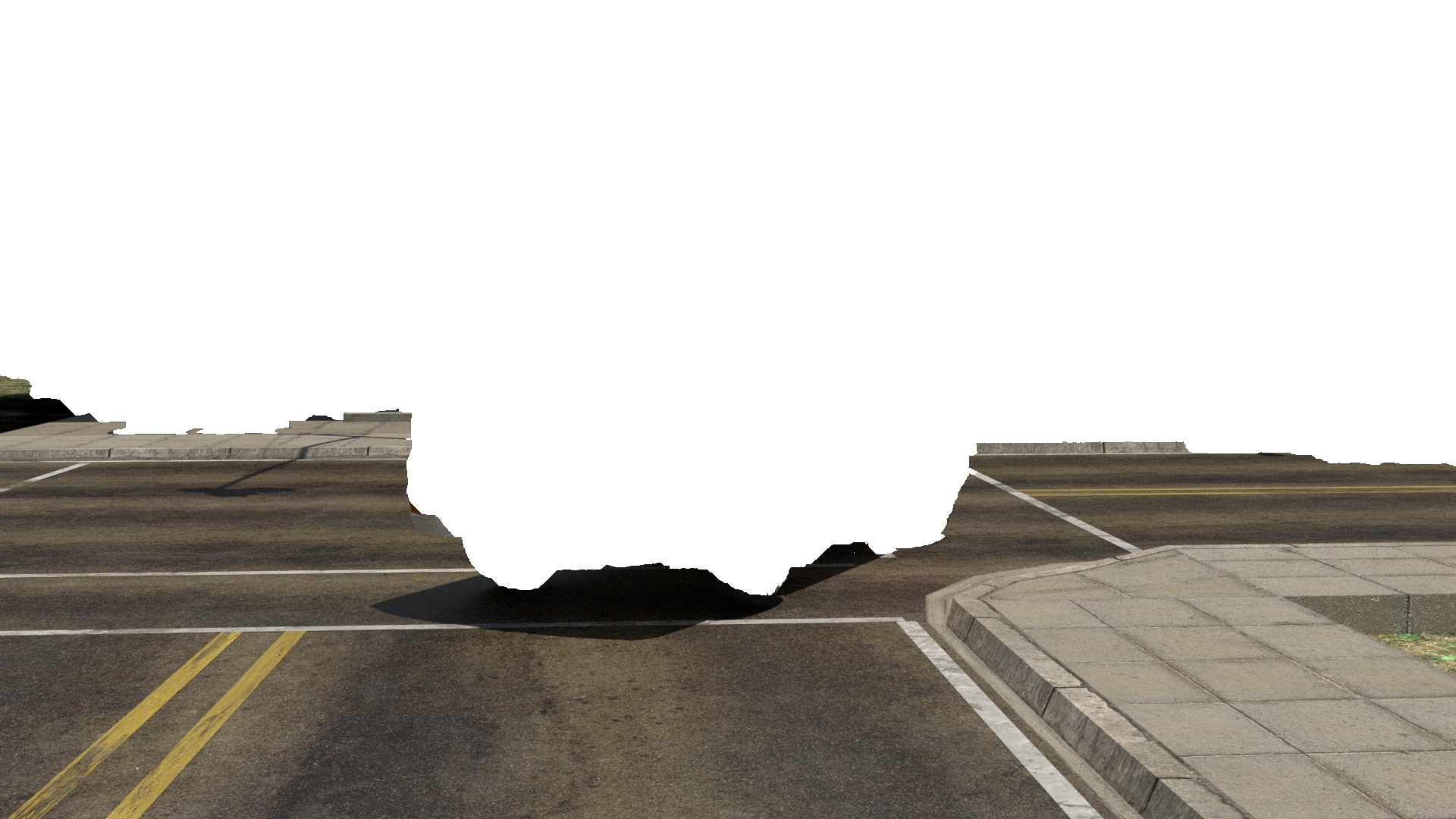}};
		\node[imageDescriptionStyle] at (sSegmentationToGroundImage.center) {Ground Segmentations};
		
		\draw (dummySfMLevelNodeRight) -- (groundComputation);
		
		\draw (objectSfM) -- (trajectoryFamilyComputation);
		\node[imageStyle] (oSfmToFamImage) at ($(objectSfM)!0.5!(trajectoryFamilyComputation) + (0,\overviewVerticalShiftSecondToThirdRow)$){ 
			\includegraphics[frame, width = \includegraficsWidthOverview]{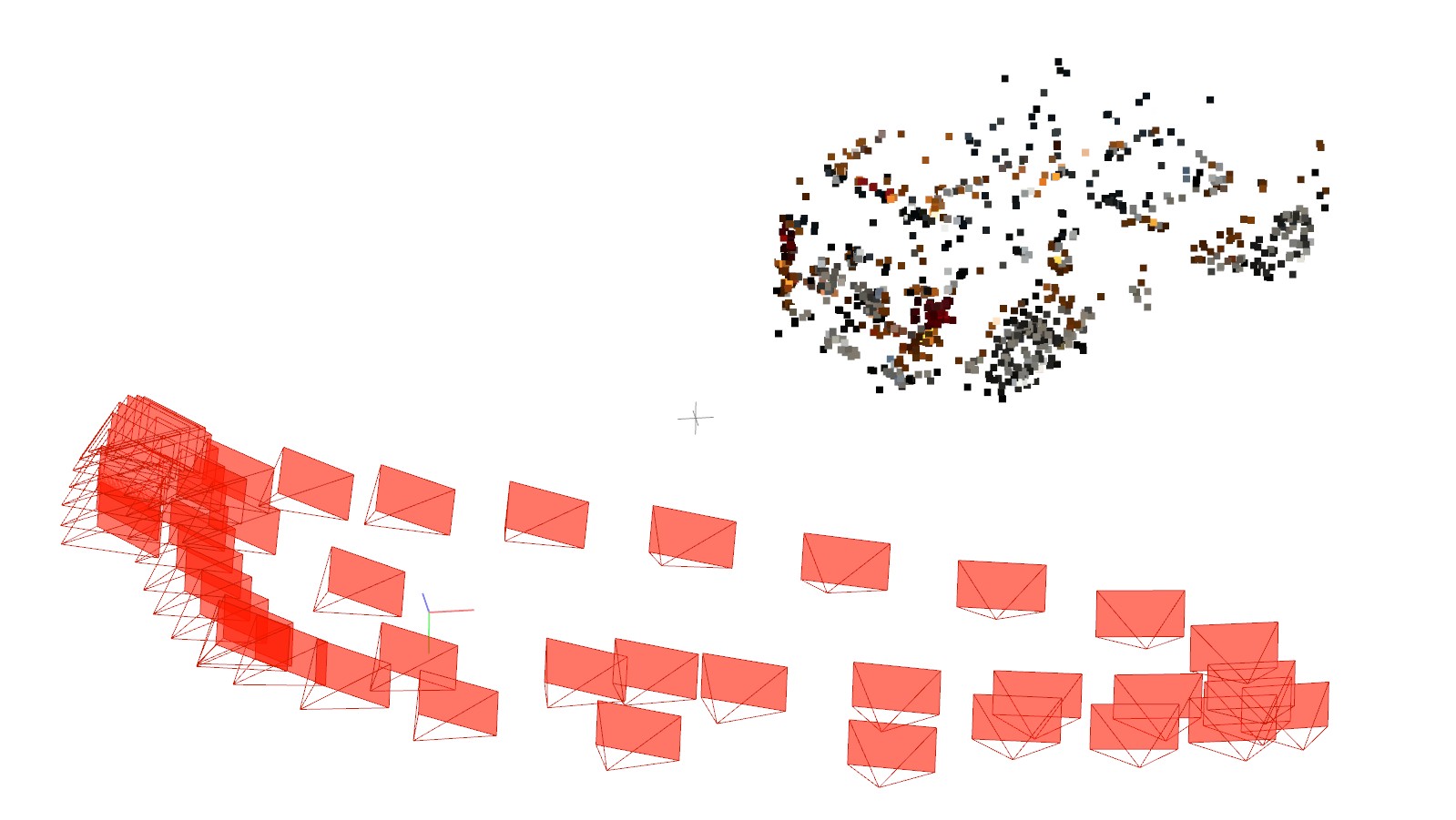}};
		\node[imageDescriptionStyle] at (oSfmToFamImage.center) {Object SfM Result};
		
		\draw (backgroundSfM) -- (trajectoryFamilyComputation);
		\node[imageStyle] (bSfmToFamImage) at ($(backgroundSfM)!0.5!(trajectoryFamilyComputation)+ (0,\overviewVerticalShiftSecondToThirdRow)$) { 
			\includegraphics[frame, width = \includegraficsWidthOverview]{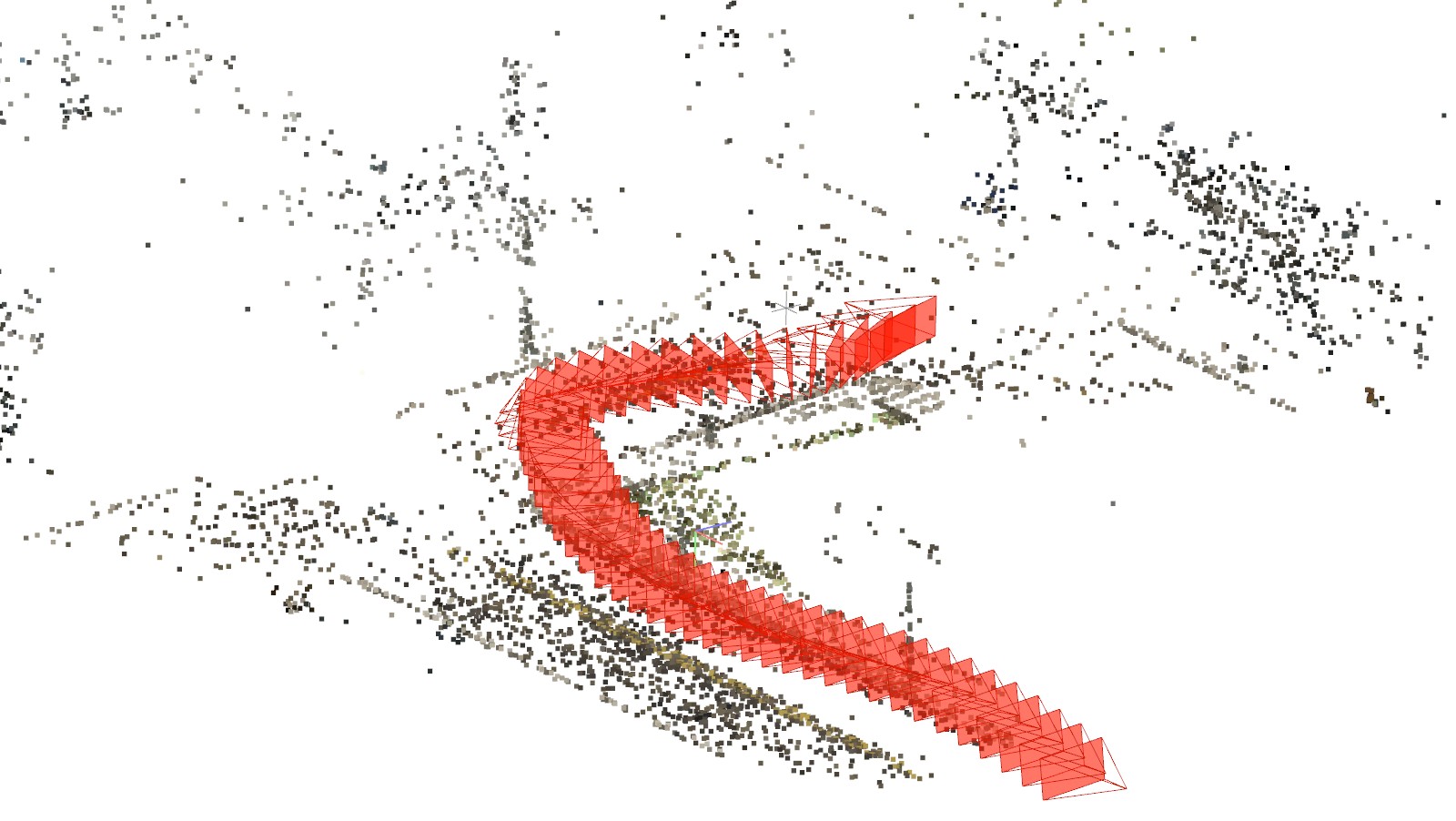}};
		\node[imageDescriptionStyle] at (bSfmToFamImage.center) {Background SfM Result};
		
		\draw (backgroundSfM) -- (groundComputation);
		\node[imageStyle]  (bSfmToGroundImage) at ($(backgroundSfM)!0.5!(groundComputation)+ (0,\overviewVerticalShiftSecondToThirdRow)$) { 
			\includegraphics[frame, width = \includegraficsWidthOverview]{background_reconstruction.jpg}};
		\node[imageDescriptionStyle] at (bSfmToGroundImage.center) {Background SfM Result};
		
		\draw (dummySfMLevelNodeRight) -- (groundComputation);
	
		\draw (trajectoryFamilyComputation) -- (scaleEstimation);
		\node[plainDescriptionStyle] at ($(trajectoryFamilyComputation)!0.5!(scaleEstimation)+(0,\overviewVerticalShiftThirdRow)$) {Object Trajectory Family};

		\draw (groundComputation) -- (scaleEstimation);
		\node[plainDescriptionStyle] at ($(groundComputation)!0.5!(scaleEstimation)+(0,\overviewVerticalShiftThirdRow)$) {Ground Representation};

		\draw (scaleEstimation) -- (finalObjectTrajectoryCoordinate) {};
	  	\node [plainDescriptionStyle] (finalObjectTrajectory) at (ConsistentObjectTrajectoryCoordinate) {Consistent Object Motion Trajectory};

	  \end{scope}

	  
	  \end{tikzpicture}
	
	
	
	
	
	\caption{Overview of the Trajectory Reconstruction Pipeline.}
	\label{methods:overview}
\end{figure*}

The pipeline of our approach is shown in Fig.~\ref{methods:overview}. The input is an ordered image sequence. We track two-dimensional object shapes on pixel level across video sequences following the approach presented in \cite{Bullinger2017}. In contrast to \cite{Bullinger2017}, we identify object shapes exploiting the instance-aware semantic segmentation method presented in \cite{Li2016} and associate extracted object shapes of subsequent frames using the optical flow approach described in \cite{Ilg2017}. Without the loss of generality, we describe motion trajectory reconstructions of single objects. We apply SfM \cite{Moulon2013,Schoenberger2016sfm} to object and background images as shown in Fig.~\ref{methods:overview}. Object images denote images containing only color information of single object instance. Similarly, background images show only background structures. We combine object and background reconstructions to determine possible, visually identical, object motion trajectories. We compute a consistent object motion trajectory exploiting constraints derived from reconstructed terrain ground geometry.

\subsection{Object Trajectory Representation}
\label{subsection:trajectory_representation}

In order to estimate a consistent object motion trajectory we apply SfM simultaneously to object and background images as shown in Fig.~\ref{methods:overview}.  We denote the corresponding SfM results with $sfm^{(o)}$ and $sfm^{(b)}$. Let $\vec{o}_{j}^{(o)} \in \mathcal{P}^{(o)}$ and $\vec{b}_{k}^{(b)} \in \mathcal{P}^{(b)}$ denote the 3D points contained in $sfm^{(o)}$ or $sfm^{(b)}$, respectively. The superscripts $o$ and $b$ in  $\vec{o}_{j}^{(o)}$ and $\vec{b}_{k}^{(b)}$ describe the corresponding coordinate frame. The variables $j$ and $k$ are the indices of points in the object or the background point cloud, respectively. We denote the reconstructed intrinsic and extrinsic parameters of each registered input image as virtual camera. Each virtual camera in $sfm^{(o)}$ and $sfm^{(b)}$ corresponds to a certain frame from which object and background images are extracted. We associate virtual cameras in $sfm^{(o)}$ with the corresponding virtual cameras in $sfm^{(b)}$ and vice versa. In the following, we consider only camera pairs, whose virtual cameras are contained in $sfm^{(o)}$ and $sfm^{(b)}$. Because of missing image registrations this may not be the case for all virtual cameras. \\
We reconstruct the object motion trajectory by combining information of corresponding virtual cameras. For any virtual camera pair of an image with index $i$ the object SfM result $sfm^{(o)}$ contains information of object point positions $\vec{o}_{j}^{(o)}$ relative to virtual cameras with camera centers $\vec{c}_{i}^{(o)}$ and rotations $\vec{R}_{i}^{(o)}$. We express each object point $\vec{o}_{j}^{(o)}$ in camera coordinates $\vec{o}_{j}^{(i)}$ of camera $i$ using equation \eqref{eq:object_to_camera_coordinates}
\begin{equation}
\label{eq:object_to_camera_coordinates}
\vec{o}_{j}^{(i)} = \vec{R}_{i}^{(o)} \cdot (\vec{o}_{j}^{(o)} - \vec{c}_{i}^{(o)}).
\end{equation}
The background SfM result $sfm^{(b)}$ contains the camera center $\vec{c}^{(b)}_{i}$ and the corresponding rotation $\vec{R}_{i}^{(b)}$, which provide pose information of the camera with respect to the reconstructed background. Note, that the camera coordinate systems of virtual cameras in $sfm^{(o)}$ and $sfm^{(b)}$ are equivalent. We use $\vec{c}_{i}^{(b)}$ and $\vec{R}_{i}^{(b)}$ to transform object points to the background coordinate system using equation \eqref{eq:camera_to_background_coordinates}
\begin{equation}
\label{eq:camera_to_background_coordinates}
\vec{o}_{j,i}^{(b)} = \vec{c}_{i}^{(b)} + \vec{R}_{i}^{T(b)} \cdot \vec{o}_{j}^{(i)}.
\end{equation}
In general, the scale ratio of object and background reconstruction does not match due to the scale ambiguity of SfM reconstructions \cite{Hartley2004}. We tackle this problem by treating the scale of the background as reference scale and by introducing a scale ratio factor $r$ to adjust the scale of object point coordinates. The overall transformation of object points given in object coordinates $\vec{o}_{j}^{(o)}$ to object points in the background coordinate frame system $\vec{o}_{j,i}^{(b)}$ of camera $i$ is described according to equation \eqref{eq:object_to_background_coordinates}.
\begin{equation}
\label{eq:object_to_background_coordinates}
\begin{split}
\vec{o}_{j,i}^{(b)} = \vec{c}_{i}^{(b)} + r \cdot \vec{R}_{i}^{T(b)} \cdot \vec{R}_{i}^{(o)} \cdot (\vec{o}_{j}^{(o)} - \vec{c}_{i}^{(o)})
\\
 \coloneqq \vec{c}_{i}^{(b)} + r \cdot \vec{v}_{j,i}^{(b)}
\end{split}
\end{equation}
with 
\begin{equation}
\label{eq:camera_to_object_point}
\vec{v}_{j,i}^{(b)} = \vec{R}_{i}^{T(b)} \cdot \vec{R}_{i}^{(o)} \cdot (\vec{o}_{j}^{(o)} - \vec{c}_{i}^{(o)}) = \vec{o}_{j,i}^{(b)} - \vec{c}_{i}^{(b)}. 
\end{equation}
Given the scale ratio $r$, we can recover the full object motion trajectory computing equation \eqref{eq:camera_to_object_point} for each virtual camera pair. We use $\vec{o}_{j,i}^{(b)}$ of all cameras and object points as object motion trajectory representation. The ambiguity mentioned in section \ref{section:introduction} is expressed by the unknown scale ratio $r$.

\subsection{Terrain Ground Approximation}
\label{subsection:ground_computation}

Further camera or object motion constraints are required to determine the scale ratio $r$ introduced in equation \eqref{eq:camera_to_object_point}. In contrast to previous work \cite{Ozden2004a,Yuan2006,Park2015,Lee2015,Song2016,Chhaya2016} we assume that the object category of interest moves on top of the terrain. We exploit semantic segmentation techniques to estimate an approximation of the ground surface of the scene. We apply the ConvNet presented in \cite{Shelhamer2017} to determine ground categories like street or grass for all input images on pixel level. We consider only stable background points, i.e. 3D points that are observed at least four times. We determine for each 3D point a ground or non-ground label by accumulating the semantic labels of corresponding keypoint measurement pixel positions. This allows us to determine a subset of background points, which represent the ground of the scene. We approximate the ground surface locally using plane representations. For each frame $i$ we use corresponding estimated camera parameters and object point observations to determine a set of ground points $P_i$ close to the object. We build a kd-tree containing all ground measurement positions of the current frame. For each object point observation, we determine the $num_{b}$ closest background measurements. In our experiments, we set $num_{b}$ to 50. Let $card_i$ be the cardinality of $P_i$. While $card_i$ is less than $num_{b}$, we add the next background observation of each point measurement. This results in an equal distribution of local ground points around the vehicle. We apply RANSAC \cite{Fischler1981} to compute a local approximation of the ground surface using $P_i$. Each plane is defined by a corresponding normal vector $\vec{n}_i$ and an arbitrary point $\vec{p}_i$ lying on the plane. 

\subsection{Scale Estimation using Constant Distance Constraints}
\label{subsection:scale_estimation_using_constant_distance}

In section \ref{subsection:scale_estimation_using_constant_distance}, we exploit priors of object motion to improve the robustness of the reconstructed object trajectory. We assume that the object of interest moves on a locally planar surface. In this case the distance of each object point $\vec{o}_{j,i}^{(b)}$ to the ground is constant for all cameras $i$. The reconstructed trajectory shows this property only for the true scale ratio and non-degenerated camera motion. For example, a
degenerate case occurs when the camera moves exactly parallel to
a planar object motion. For a more detailed discussion of degenerated camera motions see \cite{Ozden2004a}. 

\subsubsection{Scale Ratio Estimation using a Single View Pair}
We use the term \textit{view} to denote cameras and corresponding local ground planes. The signed distance of an object point $\vec{o}_{j,i}^{(b)}$ to the ground plane can be computed according to equation \eqref{eq:ground_plane_distance}
\begin{equation}
\label{eq:ground_plane_distance}
d_{j,i} = \vec{n}_{i} \cdot (\vec{o}_{j,i}^{(b)} - \vec{p}_{i}),
\end{equation}
where $\vec{p_i}$ is a point on the local ground plane and $\vec{n_i}$ is the corresponding normal vector. If the object moves on top of the approximated terrain ground the distance $d_{j,i}$ should be independent of a specific camera $i$. We substitute $d_{j,i}$ with $d_{j}$ in equation \eqref{eq:ground_plane_distance}.
This allows us to combine equation \eqref{eq:ground_plane_distance} of the same point and different cameras.
\begin{equation}
\label{eq:determine_s1}
\vec{n}_{i} \cdot (\vec{o}_{j,i}^{(b)} - \vec{p}_{i}) = \vec{n}_{i'} \cdot (\vec{o}_{j,i'}^{(b)} - \vec{p}_{i'}).
\end{equation}
Substituting equation \eqref{eq:object_to_background_coordinates} in equation \eqref{eq:determine_s1} results in \eqref{eq:determine_s2}
\begin{equation}
\label{eq:determine_s2}
\vec{n}_{i} \cdot (\vec{c}_{i}^{(b)} + r \cdot \vec{v}_{j,i}^{(b)} - \vec{p}_{i}) = \vec{n}_{i'} \cdot (\vec{c}_{i'}^{(b)} + r \cdot \vec{v}_{j,i'}^{(b)} - \vec{p}_{i'}) 
\end{equation}
Solving equation \eqref{eq:determine_s2} for $r$ yields equation \eqref{eq:determine_s3}
\begin{equation}
\label{eq:determine_s3}
	r  =  \frac{ \vec{n}_{i'} \cdot  (\vec{c}_{i'}^{(b)} - \vec{p}_{i'}) - \vec{n}_{i} \cdot (\vec{c}_{i}^{(b)} - \vec{p}_{i})}
	{(\vec{n}_{i} \cdot \vec{v}_{j,i}^{(b)} - \vec{n}_{i'} \cdot \vec{v}_{j,i'}^{(b)})}.
\end{equation}
Equation \eqref{eq:determine_s3} allows us to determine the scale ratio $r$ between object and background reconstruction using the extrinsic parameters of two cameras and corresponding ground approximations. 

\subsubsection{Scale Ratio Estimation using View Pair Ranking}
The accuracy of the estimated scale ratio $r$ in equation \eqref{eq:determine_s3} is subject to the condition of the parameters of the particular view pair. For instance, if the numerator or denominator is close to zero, small errors in the camera poses or ground approximations may result in negative scale ratios. In addition, wrongly estimated local plane normal vectors may disturb camera-plane distances. We tackle these problems by combining two different view pair rankings. The first ranking uses for each view pair the difference of the camera-plane distances, i.e. the numerator in equation \eqref{eq:determine_s3}. The second ranking reflects the quality of the local ground approximation w.r.t. the object reconstruction. For a view pair with well reconstructed local planes the variance of the corresponding scale ratios is small. This allows for the determination of ill conditioned view pairs. The second ranking uses the scale ratio difference to order the view pairs. We sort the view pairs by weighting both ranks equally. Let $vp$ denote the view pair with the lowest overall rank. The final scale ratio is determined by using a least squares method w.r.t. all equations of $vp$.




\section{Virtual Object Motion Trajectory Dataset}
\label{section:trajectory_dataset}

To quantitatively evaluate the quality of the reconstructed object motion trajectory we require accurate object and environment models as well as object and camera poses at each time step. The simultaneous capturing of corresponding ground truth data with sufficient quality is difficult to achieve. For example, one could capture the environment geometry with LIDAR sensors and the camera / object pose with an additional system. However, the registration and synchronization of all these different modalities is a complex and cumbersome process. The result will contain noise and other artifacts like drift. To tackle these issues we exploit virtual models. Previously published virtually generated and virtually augmented datasets, like \cite{Richter2016,Ros2016,Gaidon2016,Tsirikoglou2017}, provide data for different application domains and do not include three-dimensional ground truth information. We build a virtual world including an urban environment, animated vehicles as well as predefined vehicle and camera motion trajectories. This allows us to compute spatial and temporal error free ground truth data. We exploit procedural generation of textures to avoid artificial repetitions. Thus, our dataset is suitable for evaluating SfM algorithms.

\begin{figure*}
	\begin{subfigure}[t]{\subfigureWidthFourColumns}
		\includegraphics[width = \subfigureWidthFourColumns,frame]{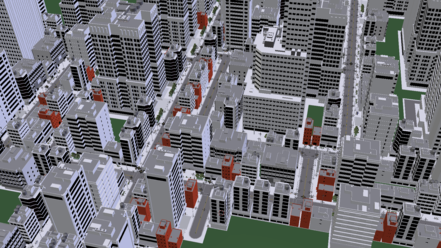}
		\caption{Environment Model from a Bird's Eye Perspective (in Blender)}
	\end{subfigure}
	\hfill
	\begin{subfigure}[t]{\subfigureWidthFourColumns}
		\includegraphics[width = \subfigureWidthFourColumns,frame]{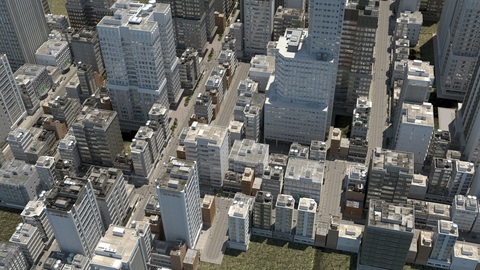}
		\caption{Environment Rendered from a Bird's Eye Perspective (Rendered)}
	\end{subfigure}
	\hfill
	\begin{subfigure}[t]{\subfigureWidthFourColumns}
		\includegraphics[width = \subfigureWidthFourColumns,frame]{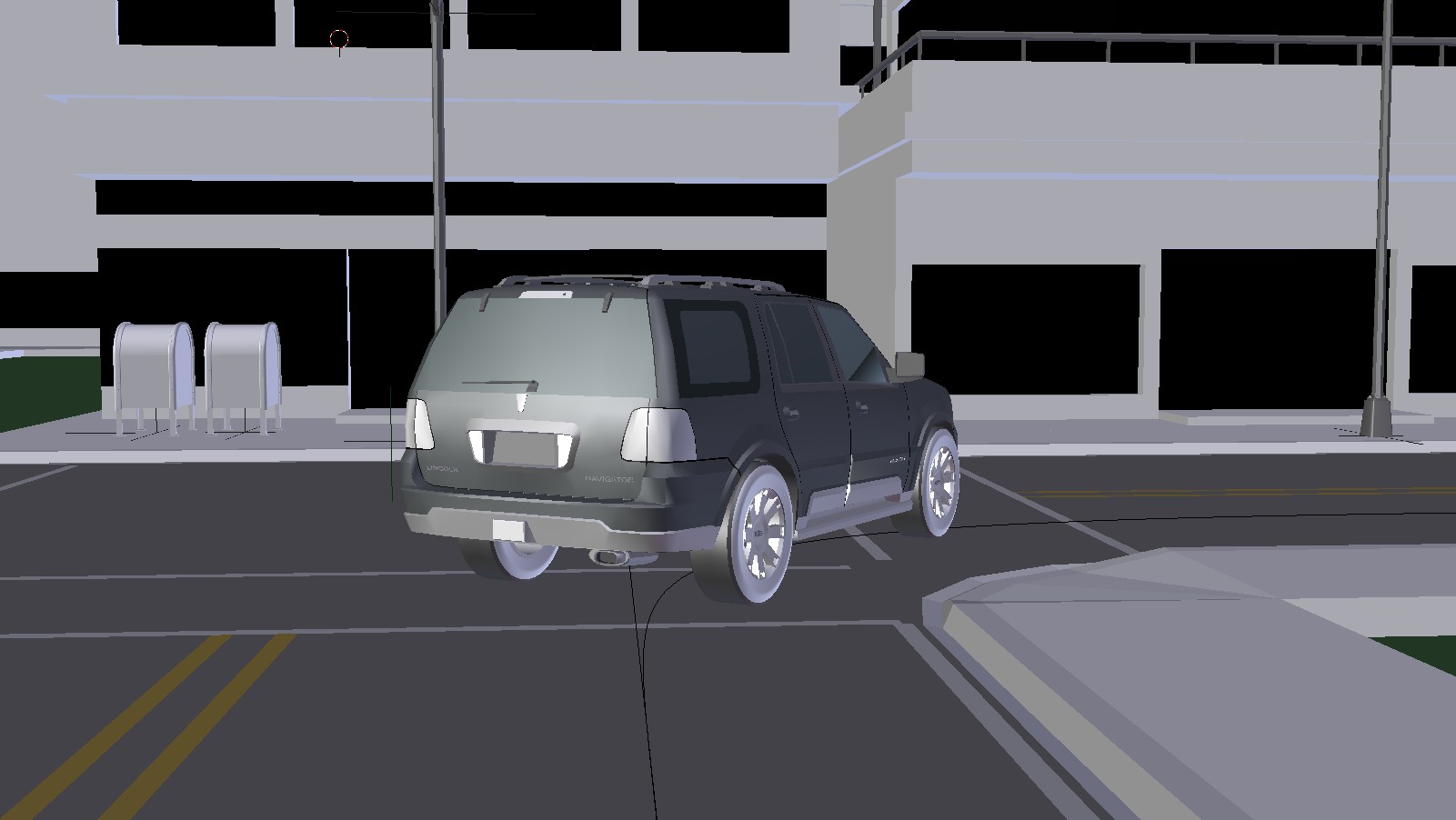}
		\caption{Car Model with Motion Path in Street Scene (in Blender)}
	\end{subfigure}
	\hfill
	\begin{subfigure}[t]{\subfigureWidthFourColumns}
		\includegraphics[width = \subfigureWidthFourColumns,frame]{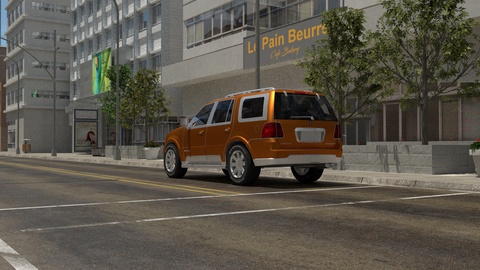}
		\caption{Rendered Street Scene}
	\end{subfigure}
	\caption{Example Scenes from the Virtual Object Trajectory Dataset.}
	\label{fig:evaluation:dataset_scenes}
\end{figure*}

\begin{figure*}
	\begin{subfigure}[t]{\subfigureWidthFourColumns}
		\includegraphics[width=\subfigureWidthFourColumns,frame]{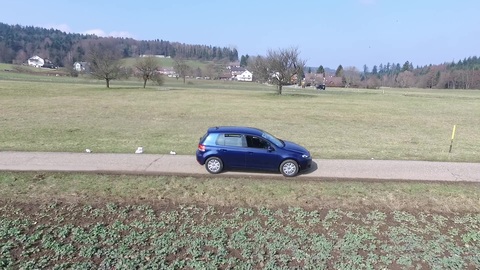}
		\caption{Input Frame 0}
		\label{fig:input_frame_0}
	\end{subfigure}
	\hfill
	\begin{subfigure}[t]{\subfigureWidthFourColumns}
		\includegraphics[width = \subfigureWidthFourColumns,frame]{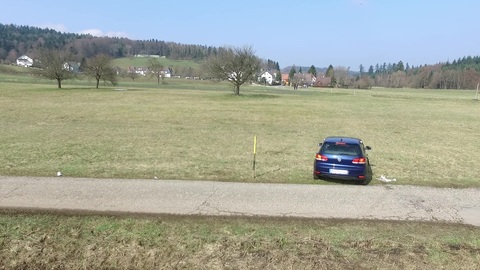}
		\caption{Input Frame 100}
		\label{fig:input_frame_100}
	\end{subfigure}
	\hfill
	\begin{subfigure}[t]{\subfigureWidthFourColumns}
		\includegraphics[width = \subfigureWidthFourColumns,frame]{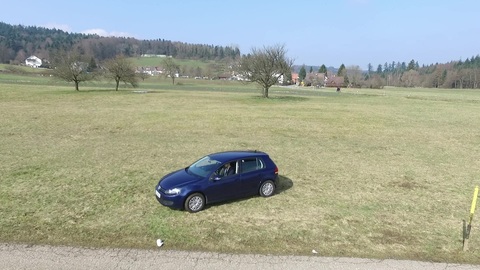}
		\caption{Input Frame 200}
		\label{fig:input_frame_200}
	\end{subfigure}
	\hfill
	\begin{subfigure}[t]{\subfigureWidthFourColumns}
		\includegraphics[width = \subfigureWidthFourColumns,frame]{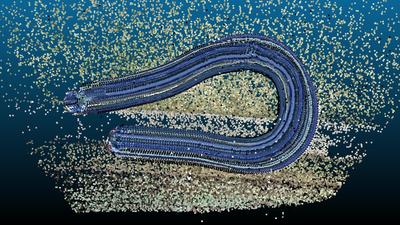}
		\caption{Reconstructed Trajectory (Top View)}
		\label{fig:real_trajectory_top_view}
	\end{subfigure}
	\caption{Car Trajectory Reconstruction using 250 frames with a resolution of 1920 x 1080 pixels captured by a DJI drone.}
	\label{fig:trajectory_reconstruction_real}
\end{figure*}

\begin{figure*}
	\begin{subfigure}[t]{\subfigureWidthFourColumns}
		\includegraphics[width = \subfigureWidthThreeColumns,frame]{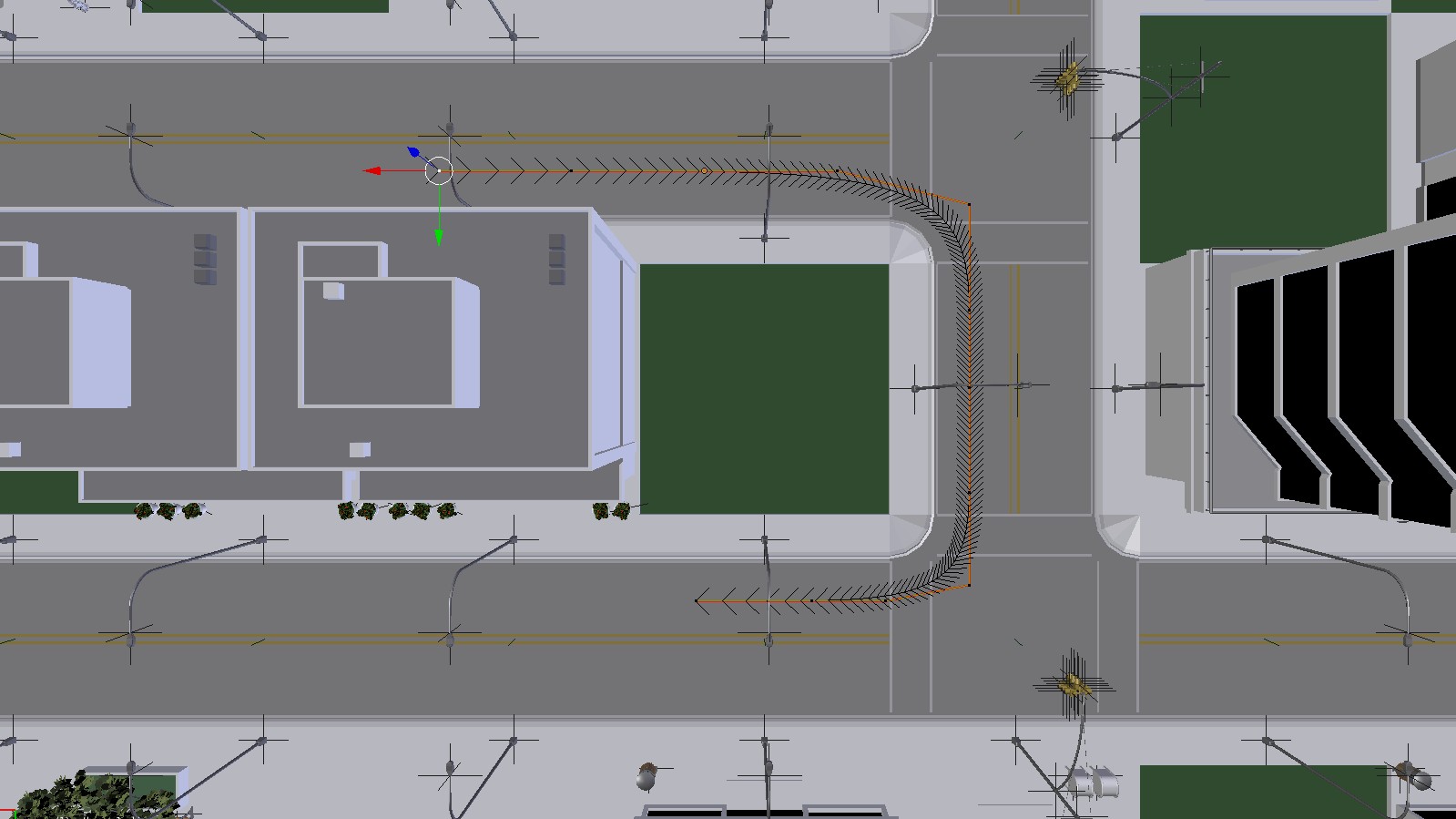}
		\caption{Ground Truth Object Path}
		\label{fig:evaluation:quantitative_evaluation_object_path}
	\end{subfigure}
	\begin{subfigure}[t]{\subfigureWidthFourColumns}
		\includegraphics[width = \subfigureWidthThreeColumns,frame]{model_in_scene.jpg}
		\caption{Object Model in Blender}
		\label{fig:evaluation:quantitative_evaluation_object_model}
	\end{subfigure}
	\begin{subfigure}[t]{\subfigureWidthFourColumns}
		\includegraphics[width = \subfigureWidthThreeColumns,frame]{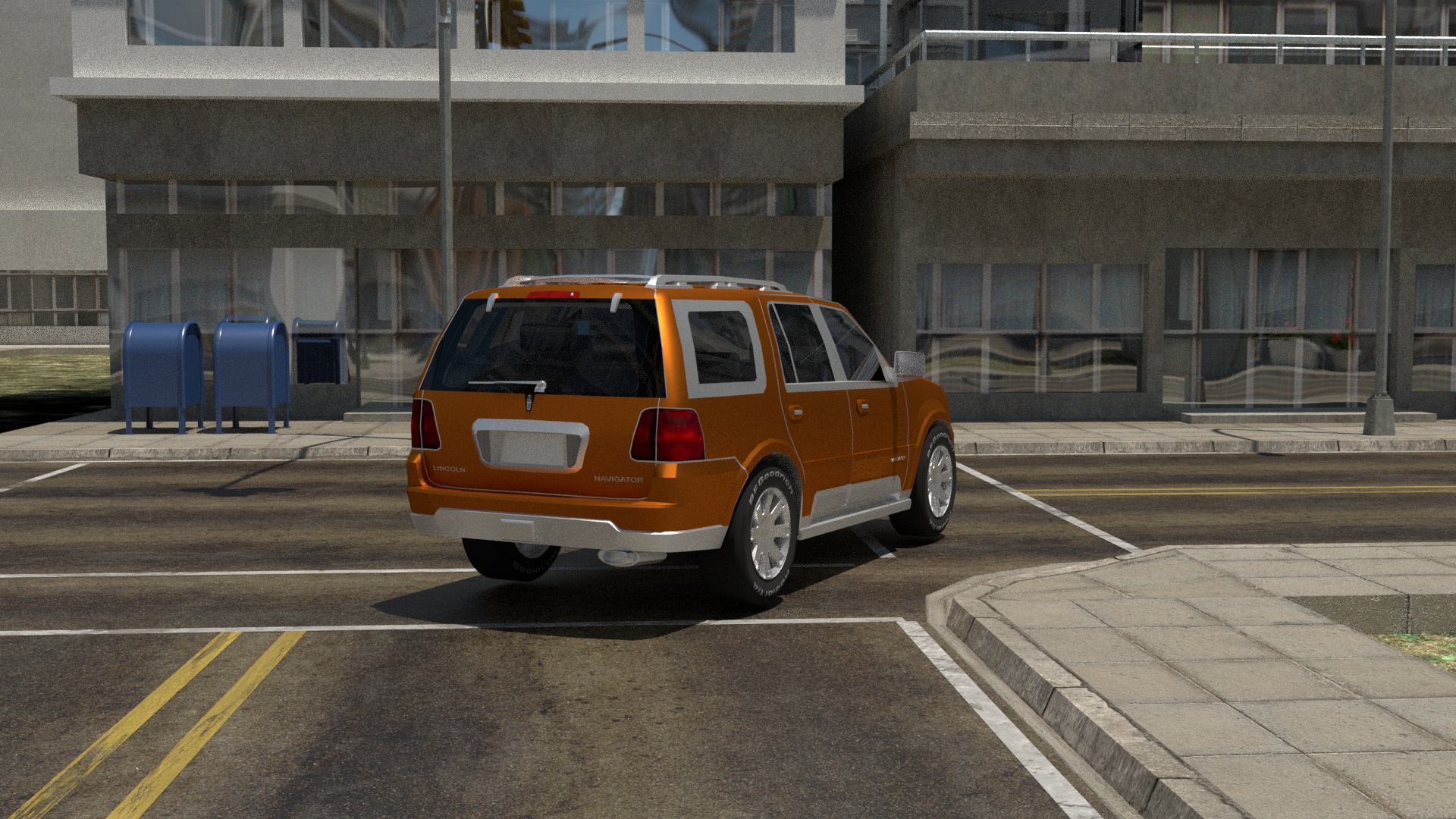}
		\caption{Rendered Scene}
		\label{fig:evaluation:quantitative_evaluation_rendered_scene}
	\end{subfigure}
	\begin{subfigure}[t]{\subfigureWidthFourColumns}
		\includegraphics[width = \subfigureWidthThreeColumns,frame]{frame00039_object_seg.jpg}
		\caption{Object Segmentation}
		\label{fig:evaluation:quantitative_evaluation_object_seg}
	\end{subfigure}
	\begin{subfigure}[t]{\subfigureWidthFourColumns}
		\includegraphics[width = \subfigureWidthThreeColumns,frame]{frame00039_background_seg.jpg}
		\caption{Background Segmentation}
		\label{fig:evaluation:quantitative_evaluation_background_sec}
	\end{subfigure}
	\begin{subfigure}[t]{\subfigureWidthFourColumns}
		\includegraphics[width = \subfigureWidthThreeColumns,frame]{frame00039_ground_seg.jpg}
		\caption{Ground Segmentation}
		\label{fig:evaluation:quantitative_evaluation_ground_sec}
	\end{subfigure}
	\begin{subfigure}[t]{\subfigureWidthFourColumns}
		\includegraphics[width = \subfigureWidthThreeColumns,frame]{object_reconstruction.jpg}
		\caption{Object Reconstruction}
		\label{fig:evaluation:quantitative_evaluation_obj_rec}
	\end{subfigure}
	\begin{subfigure}[t]{\subfigureWidthFourColumns}
		\includegraphics[width = \subfigureWidthThreeColumns,frame]{background_reconstruction.jpg}
		\caption{Background Reconstruction}
		\label{fig:evaluation:quantitative_evaluation_background_rec}
	\end{subfigure}
	\begin{subfigure}[t]{\subfigureWidthFourColumns}
		\includegraphics[width = \subfigureWidthThreeColumns,frame]{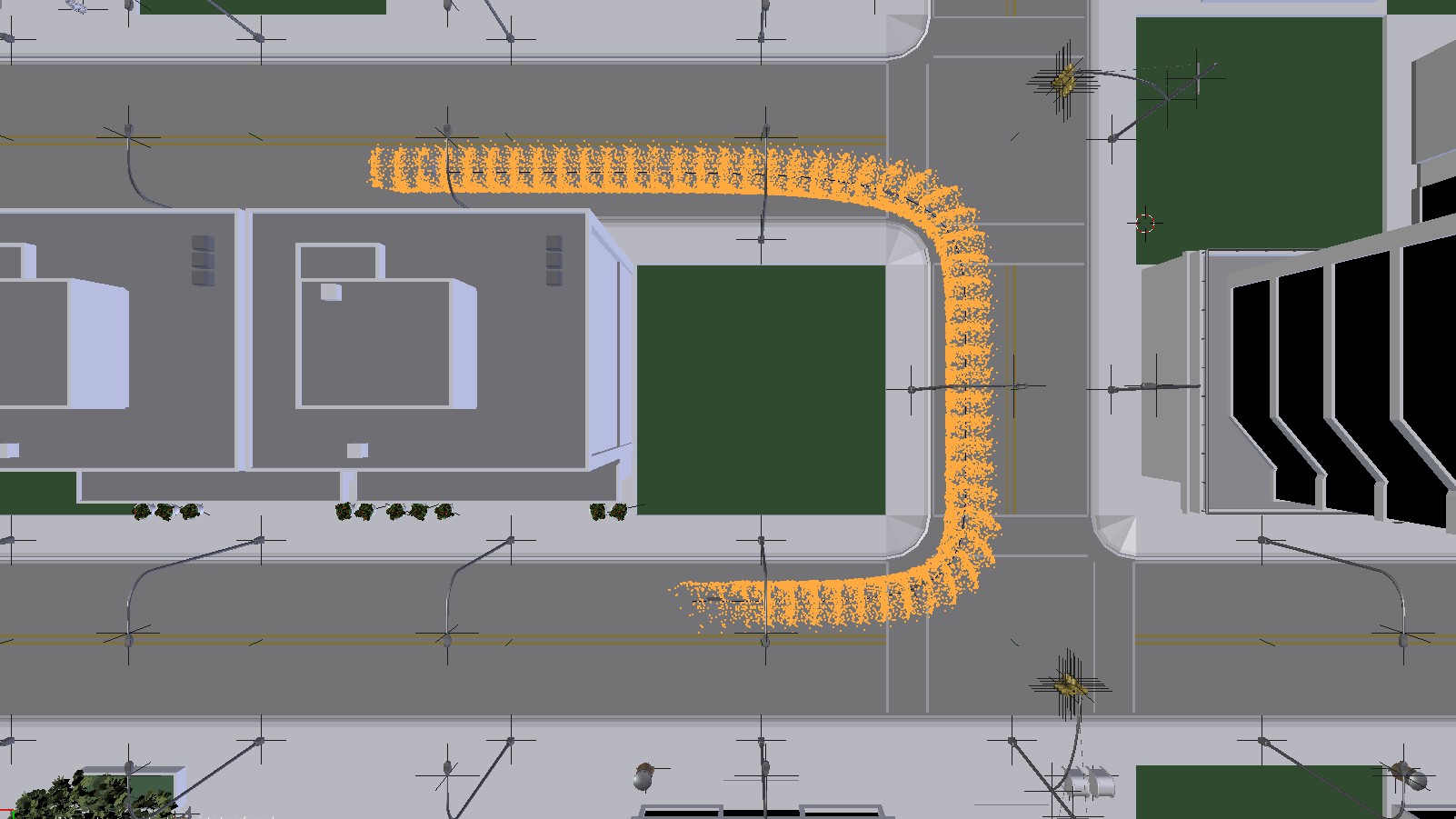}
		\caption{Reconstructed Object Motion Trajectory}
		\label{fig:evaluation:quantitative_evaluation_overlay_top}
	\end{subfigure}
	\hfill
	\begin{subfigure}[t]{\subfigureWidthFourColumns}
		\includegraphics[width = \subfigureWidthThreeColumns,frame]{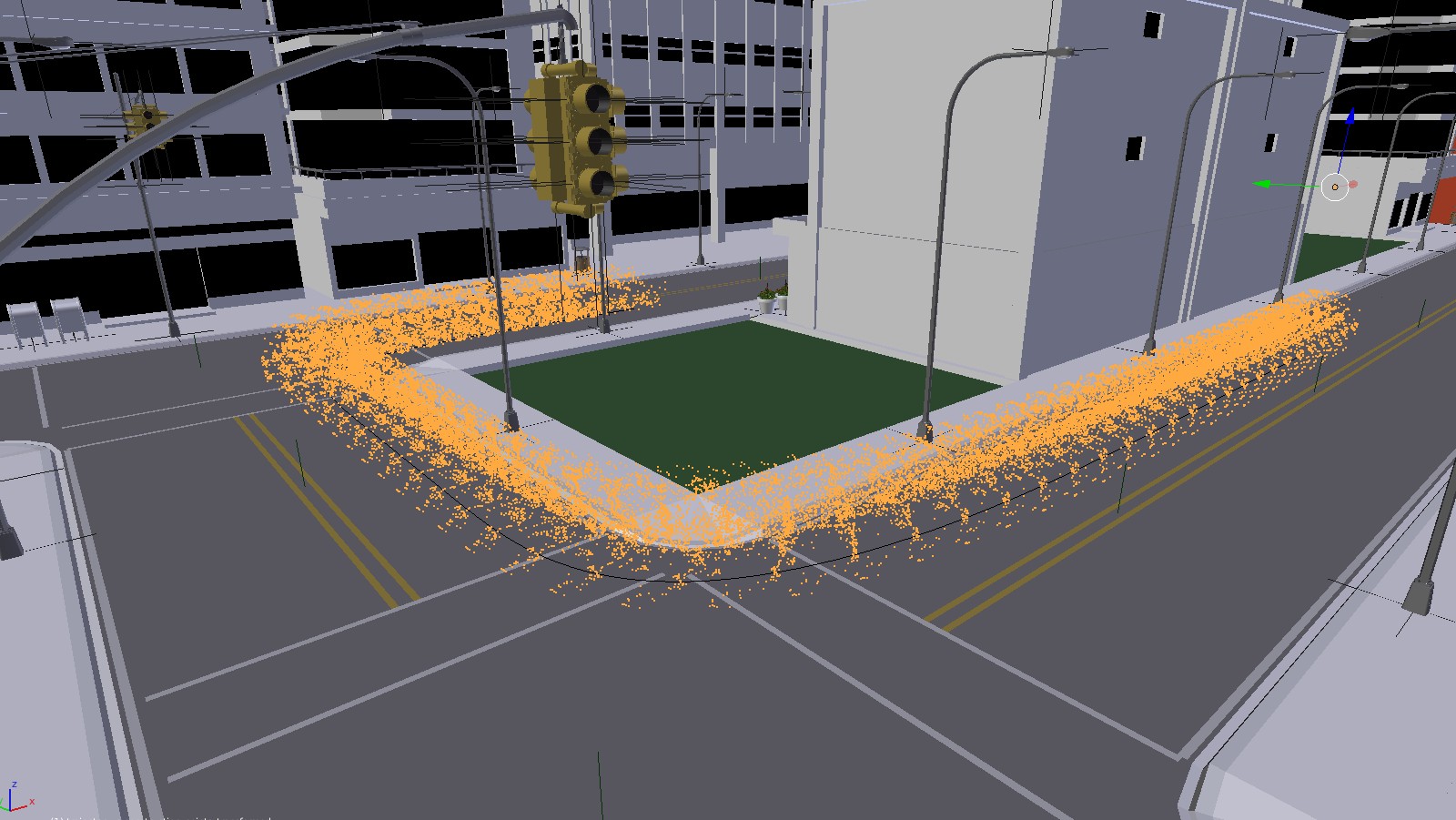}
		\caption{Reconstructed Object Motion Trajectory}
		\label{fig:evaluation:quantitative_evaluation_overlay_side}
	\end{subfigure}
	\hfill
	\begin{subfigure}[t]{\subfigureWidthFourColumns}
		\includegraphics[width = \subfigureWidthThreeColumns,frame]{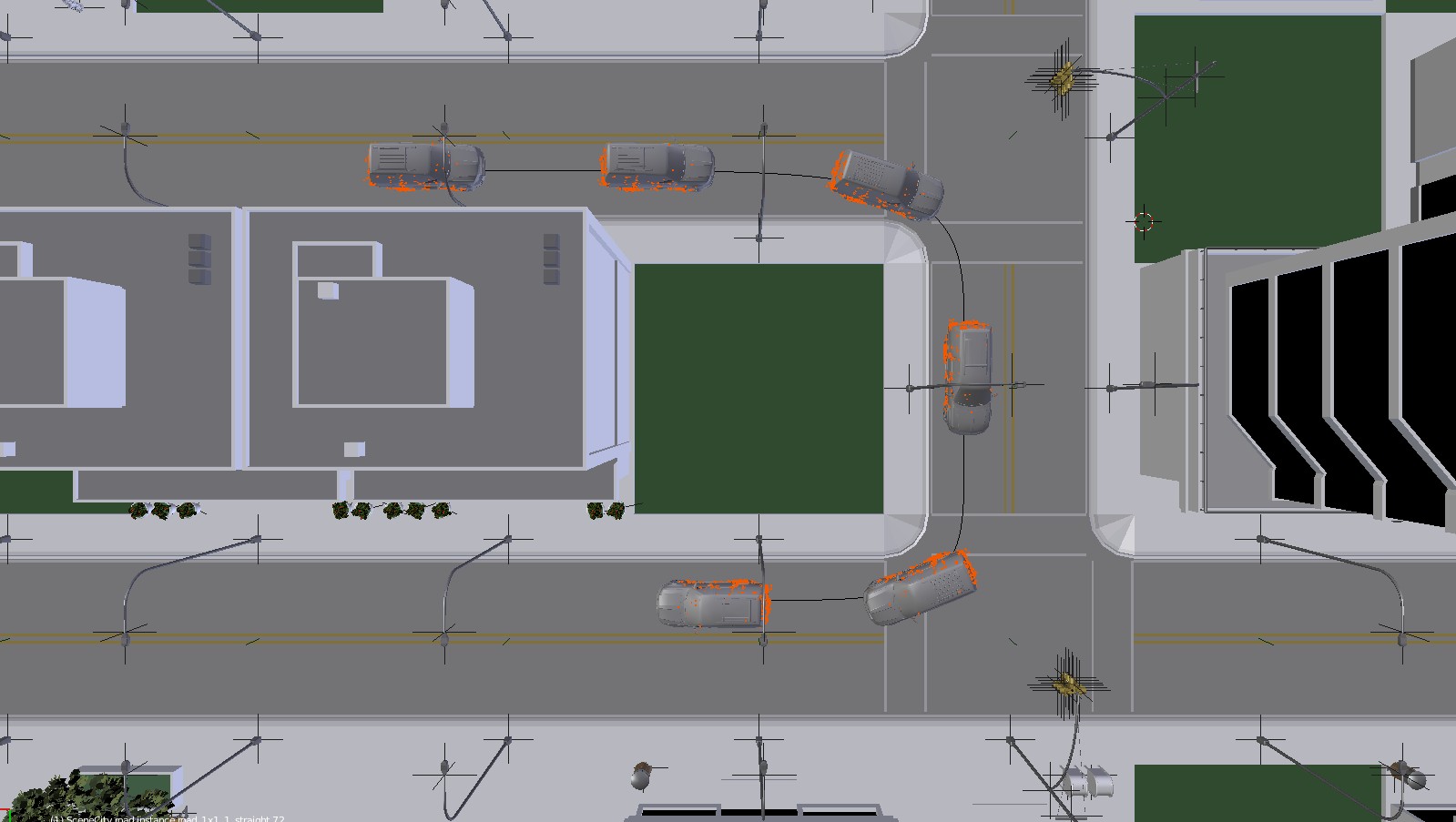}
		\caption{Ground Truth Model at Different Frames Overlayed With Reconstruction}
		\label{fig:evaluation:quantitative_evaluation_ground_truth_with_points_1}
	\end{subfigure}
	\hfill
	\begin{subfigure}[t]{\subfigureWidthFourColumns}
		\includegraphics[width = \subfigureWidthThreeColumns,frame]{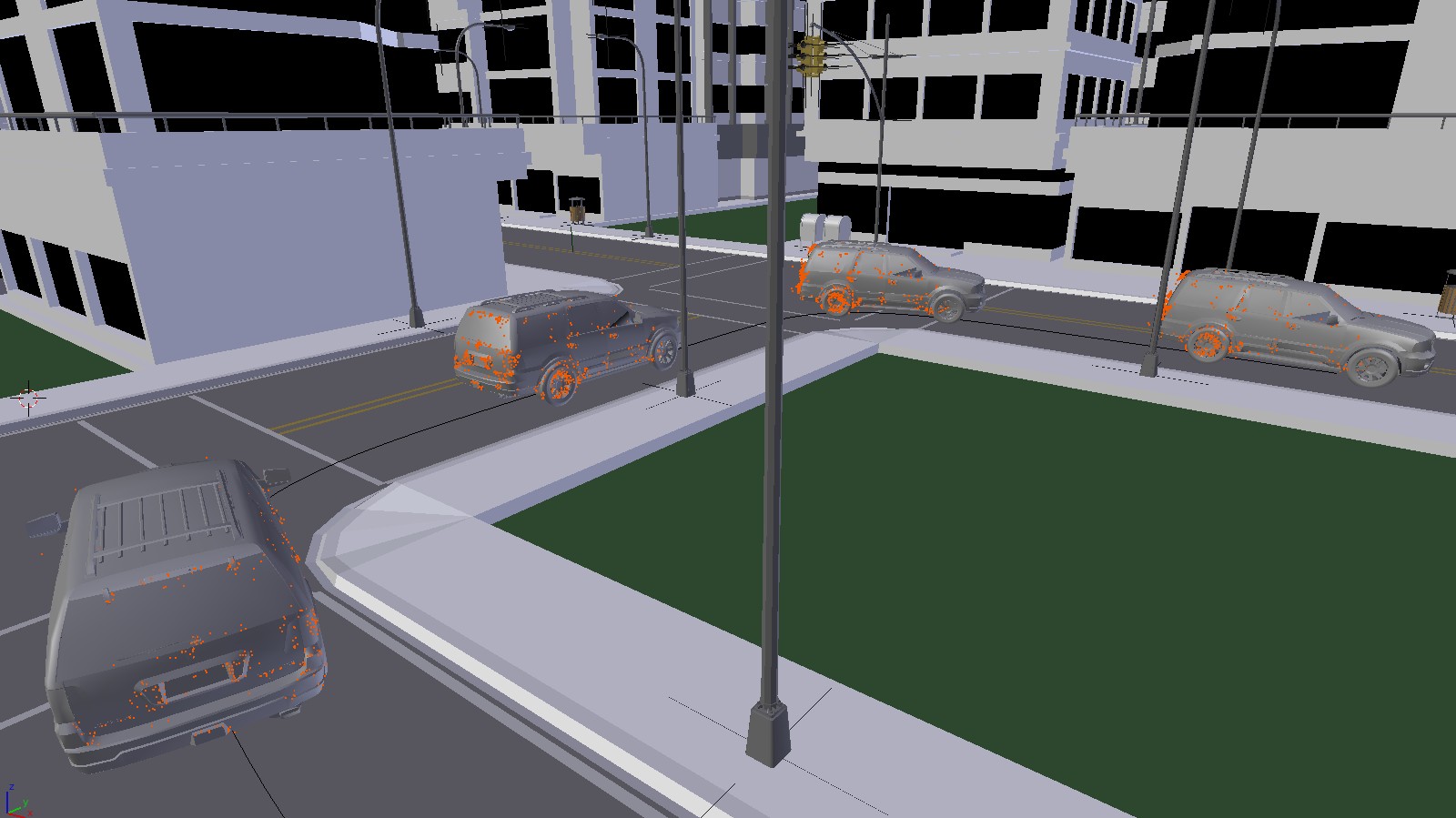}
		\caption{Ground Truth Model at Different Frames Overlayed With Reconstruction}
		\label{fig:evaluation:quantitative_evaluation_ground_truth_with_points_2}
	\end{subfigure}
	\caption{Car trajectory reconstruction using 60 virtual frames with a resolution of 1920 x 1080 pixels.}
	\label{fig:evaluation:quantitative_evaluation}
\end{figure*}

\subsection{Virtual World}

We used Blender \cite{Blender2016} to create a virtual world consisting of a city surrounded by a countryside. We exploit procedural generation to compute textures of large surfaces, like streets and sidewalks, to avoid degenerated Structure from Motion results caused by artificial texture repetitions. The virtual world includes different assets like trees, traffic lights, streetlights, phone booths, bus stops and benches. We collected a set of publicly available vehicle assets to populate the scenes. We used skeletal animation, also referred to as rigging, for vehicle animation. This includes wheel rotation and steering w.r.t. the motion trajectory as well as consistent vehicle placement on uneven ground surfaces. The animation of wheels is important to avoid unrealistic wheel point triangulations. We adjusted the scale of vehicles and virtual environment using Blender's unit system. This allows us to set the virtual space in relation to the real world. The extent of the generated virtual world corresponds to one square kilometer. We exploit environment mapping to achieve realistic illumination. With Blender's built-in tools, we defined a set of camera and object motion trajectories. This allows us to determine the exact 3D pose of cameras and vehicles at each time step.  

\subsection{Trajectory Dataset}
We use the previously created virtual world to build a new object trajectory dataset\textsuperscript{\ref{project_page}}. The dataset consists of 35 sequences capturing five vehicles in different urban scenes.  Fig.~\ref{fig:evaluation:dataset_scenes} shows some example images. The virtual video sequences cover a high variety of object and camera poses. The object trajectories reflect common vehicle motions include vehicle acceleration, different curve types and motion on changing slopes. We use the path-tracing render engine Cycles \cite{Blender2016} to achieve photo realistic rendering results. We observed that the removal of artificial path-tracing artifacts using denoising is crucial to avoid degenerated SfM reconstructions. \\
The dataset includes 6D object and camera poses for each frame as well as ground truth meshes of corresponding vehicle models. In contrast to measured ground truth data, virtual ground truth data is free of noise and shows no spatial registration or temporal synchronization inaccuracies. The dataset contains semantic segmentations of objects, ground and background to separate the reconstruction task from specific semantic segmentation and tracking approaches. In addition to the virtual data, the dataset also includes the computed reconstruction results. We will make our evaluation scripts publicly available to foster future analysis of object trajectory estimation.

\section{Experiments and Evaluation}
\label{section:experiments_and_evaluation}

We show qualitative results and quantitative evaluations using real drone footage and virtual generated drone imagery, respectively. Fig.~\ref{fig:trajectory_reconstruction_real} shows the reconstruction of a car motion trajectory using images captured by a DJI drone. Fig.~\ref{fig:evaluation:quantitative_evaluation} shows an example of the virtual object trajectory dataset. Fig.~\ref{fig:evaluation:quantitative_evaluation_overlay_side} and Fig.~\ref{fig:evaluation:quantitative_evaluation_overlay_top} show the object point cloud transformed into the virtual world coordinate frame system. Fig.~\ref{fig:evaluation:quantitative_evaluation_ground_truth_with_points_1} and Fig.~\ref{fig:evaluation:quantitative_evaluation_ground_truth_with_points_2} show the overlay result of transformed points and the corresponding virtual ground truth model. To segment the two-dimensional shape of objects of interest we follow the approach presented in \cite{Bullinger2017}. In contrast to \cite{Bullinger2017}, we used \cite{Li2016} and \cite{Ilg2017} to segment and track visible objects, respectively. We considered the following SfM pipelines for object and background reconstructions: Colmap \cite{Schoenberger2016sfm}, OpenMVG \cite{Moulon2013}, Theia \cite{Sweeney2014} and VisualSfM \cite{Wu2011}. Our object trajectory reconstruction pipeline uses Colmap for object and OpenMVG for background reconstructions, since Colmap and OpenMVG created in our experiments the most reliable object and background reconstructions.

\subsection{Virtual Object Motion Trajectory Evaluation}
\label{subsection:experiments_virtual_object_trajectory_evaluation}

We use the dataset presented in section \ref{section:trajectory_dataset} to quantitatively evaluate the proposed object motion trajectory reconstruction approach. The evaluation is based on object, background and ground segmentations included in the dataset. This allows us to show results independent from the performance of specific instance segmentation and tracking approaches. We compare the proposed method with the baseline presented in section \ref{subsection:scale_estimation_intersection} using 35 sequences contained in the dataset. We automatically register the reconstructed object trajectory to the ground truth using the method described in section \ref{subsection:registration}. We compute the shortest distance of each object trajectory point to the object mesh in ground truth coordinates. For each sequence we define the trajectory error as the average trajectory-point-mesh distance. Fig.~\ref{fig:evaluation:quantitative_trajectory_error} shows for each sequence the trajectory error in meter. The average trajectory error per vehicle using the full dataset is shown in table \ref{table:quantitative:evaluation_summary}. Overall, we achieve a trajectory error of 0.31 meter. The error of the object trajectory reconstructions reflects four types of computational inaccuracies: deviations of camera poses w.r.t. object and background point clouds, wrong triangulated object points as well as scale ratio discrepancies. Fig.~\ref{fig:evaluation:quantitative_scale_ratio} compares the estimated scale ratios of the proposed and the baseline method w.r.t. the reference scale ratio. The reference scale ratio computation is described in section \ref{subsection:ground_truth_scale_ratio_computation}. The overall estimated scale ratio deviation w.r.t. the reference scale per vehicle is shown in table \ref{table:quantitative:evaluation_summary}. The provided reference scale ratios are subject to the registration described in section \ref{subsection:registration}. Wrongly reconstructed background camera poses may influence the reference scale ratio. The \textit{van} object reconstruction was only partial successful on the sequences \textit{crossing}, \textit{overtaking} and \textit{steep street}. The SfM algorithm registered $19\%$, $60\%$ and $98\%$ of the images, respectively. The object reconstruction of the \textit{smart} model contained $74\%$ of the \textit{crossing} input object images. Here, we use the subset of registered images to perform the evaluation. The camera and the object motion in \textit{bumpy road} simulate a sequence close to a degenerated case, i.e. equation \eqref{eq:determine_s3} is ill conditioned for all view pairs. 





\begin{figure*}
	
	
	\centering
	
	\begin{tikzpicture}%
	\begin{axis}[
	legend columns=2,
	transpose legend,
	height=5cm,
	width=\textwidth,
	bar width=3pt,
	grid=major,
	ybar,
	symbolic x coords={Right Curves,Left Curves,Crossing,Overtaking,Bridge,Steep Street,Bumpy Road},
	xtick = data,
	ylabel=Trajectory Error in meter,
	ymax=2.5]%

\addplot[ybar,fill=red,postaction={
	pattern=north east lines
}] coordinates {%
	(Right Curves,0.053598095729)%
	(Left Curves,0.602482901735)%
	(Crossing,0.134532845449)%
	(Overtaking,1.37980092554)%
	(Bridge,0.46756546579)%
	(Steep Street,0.0429242934073)%
	(Bumpy Road,0.250726689975)%
};%
%
\addplot[ybar,fill=red] coordinates {%
	(Right Curves,0.0779823477051)%
	(Left Curves,0.294635890225)%
	(Crossing,0.16421817284)%
	(Overtaking,0.147330959434)%
	(Bridge,0.491235998101)%
	(Steep Street,0.0479360408705)%
	(Bumpy Road,0.195888378597)%
};%
%
\addplot[ybar,fill=green,postaction={
	pattern=north east lines
}] coordinates {%
	(Right Curves,0.269884544192)%
	(Left Curves,0.770079333959)%
	(Crossing,0.354087752238)%
	(Overtaking,0.573750699102)%
	(Bridge,0.670834054373)%
	(Steep Street,0.939725972772)%
	(Bumpy Road,0.155655800467)%
};%
%
\addplot[ybar,fill=green] coordinates {%
	(Right Curves,0.106319501631)%
	(Left Curves,0.0730803265384)%
	(Crossing,0.137755197904)%
	(Overtaking,0.810359739146)%
	(Bridge,0.145889167979)%
	(Steep Street,0.0905222533534)%
	(Bumpy Road,0.232464727986)%
};%
%
\addplot[ybar,fill=blue,postaction={
	pattern=north east lines
}] coordinates {%
	(Right Curves,0.253936651033)%
	(Left Curves,0.351694919067)%
	(Crossing,0.225660827137)%
	(Overtaking,0.153376482288)%
	(Bridge,0.0506197863029)%
	(Steep Street,0.209123945321)%
	(Bumpy Road,0.474221641169)%
};%
%
\addplot[ybar,fill=blue] coordinates {%
	(Right Curves,0.128201379922)%
	(Left Curves,0.177278595279)%
	(Crossing,0.0961720601452)%
	(Overtaking,1.22295458793)%
	(Bridge,0.252153352029)%
	(Steep Street,0.0658269534052)%
	(Bumpy Road,0.382326235226)%
};%
%
\addplot[ybar,fill=cyan,postaction={
	pattern=north east lines
}] coordinates {%
	(Right Curves,0.108071969058)%
	(Left Curves,0.539533762998)%
	(Crossing,0.248618298542)%
	(Overtaking,4.81176623115)%
	(Bridge,0.601825019339)%
	(Steep Street,0.242294112916)%
	(Bumpy Road,0.121167735881)%
};%
%
\addplot[ybar,fill=cyan] coordinates {%
	(Right Curves,0.285854159626)%
	(Left Curves,0.306516151327)%
	(Crossing,0.857361353992)%
	(Overtaking,0.389435727234)%
	(Bridge,0.0665042496999)%
	(Steep Street,0.183560460788)%
	(Bumpy Road,0.2402708233)%
};%
%
\addplot[ybar,fill=magenta,postaction={
	pattern=north east lines
}] coordinates {%
	(Right Curves,6.2542432696)%
	(Left Curves,4.0036930986)%
	(Crossing,0.178448101162)%
	(Overtaking,0.455767467004)%
	(Bridge,0.359445594915)%
	(Steep Street,0.236426639146)%
	(Bumpy Road,0.253784971619)%
};%
%
\addplot[ybar,fill=magenta] coordinates {%
	(Right Curves,0.630170783475)%
	(Left Curves,0.748985454125)%
	(Crossing,0.368657340165)%
	(Overtaking,0.648114507444)%
	(Bridge,0.386703981171)%
	(Steep Street,0.144823575715)%
	(Bumpy Road,0.333642818506)%
};%
%

	\legend{
	Lancer (Baseline),Lancer (Ours),
	Lincoln (B.),Lincoln (O.),
	Smart (B.),Smart (O.),
	Golf (B.),Golf (O.),
	Van (B.),Van (O.)}

	\end{axis}%
	\end{tikzpicture}%
	\caption{Quantitative evaluation of the trajectory error in meter computed by the methods \textit{constant distance} (ours) and \textit{intersection} (baseline). We evaluate seven different vehicle trajectories (\textit{right curves}, \textit{left curves}, \textit{crossing}, \textit{overtaking}, \textit{bridge}, \textit{steep street} and \textit{bumpy road}) and five different vehicle models (\textit{Lancer}, \textit{Lincoln Navigator}, \textit{Smart}, \textit{Golf} and \textit{Van}). The cropped values of the baseline are $6.3$m, $4$m and $4.8$m.}
	\label{fig:evaluation:quantitative_trajectory_error}
\end{figure*}
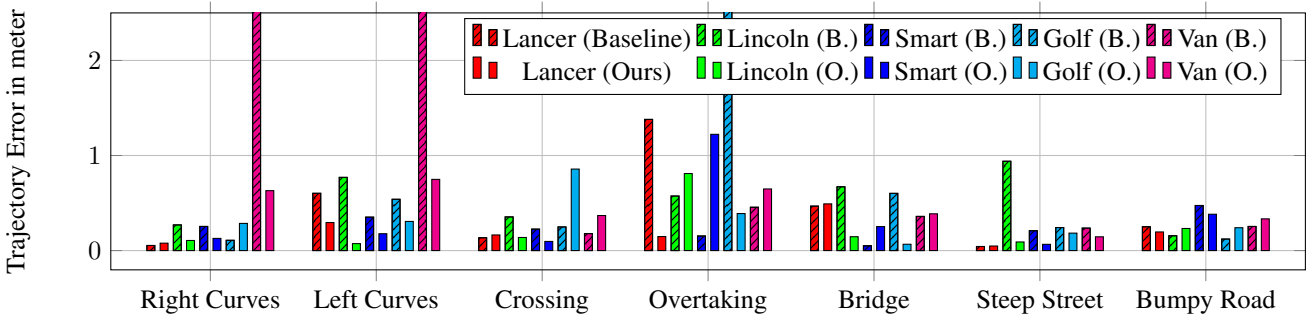
\begin{figure*}
\centering


	\begin{tikzpicture}
		\begin{axis}[
		legend columns=2,
		transpose legend,
		height=5cm,
		width=\textwidth,
		bar width=3pt,
		ybar,
		grid=major,
		symbolic x coords={Right Curves,Left Curves,Crossing,Overtaking,Bridge,Steep Street,Bumpy Road},
		xtick = data,
		ylabel=Deviation w.r.t. Reference]%

	\addplot[ybar,fill=red,postaction={
		pattern=north east lines
	}] coordinates {%
		(Right Curves,0.00117278602057)%
		(Left Curves,0.0994458597819)%
		(Crossing,0.0155003207533)%
		(Overtaking,0.157978684017)%
		(Bridge,0.0399849396201)%
		(Steep Street,0.00111344492674)%
		(Bumpy Road,0.00664351520934)%
	};%
	%
	\addplot[ybar,fill=red] coordinates {%
		(Right Curves,0.00668255517527)%
		(Left Curves,0.00601005508218)%
		(Crossing,0.0209177975149)%
		(Overtaking,0.062044353744)%
		(Bridge,0.166979654114)%
		(Steep Street,0.00237700906692)%
		(Bumpy Road,0.0225547971802)%
	};%
	%
	\addplot[ybar,fill=green,postaction={
		pattern=north east lines
	}] coordinates {%
		(Right Curves,0.0340236833872)%
		(Left Curves,0.137890116963)%
		(Crossing,0.0499464519698)%
		(Overtaking,0.0149051979015)%
		(Bridge,0.117566638912)%
		(Steep Street,0.121468240841)%
		(Bumpy Road,0.0208024796445)%
	};%
	%
	\addplot[ybar,fill=green] coordinates {%
		(Right Curves,0.00208243138544)%
		(Left Curves,0.000907929850387)%
		(Crossing,0.0131114661508)%
		(Overtaking,0.0536186787146)%
		(Bridge,0.0257749172787)%
		(Steep Street,0.0204879046834)%
		(Bumpy Road,0.157479814048)%
	};%
	%
	\addplot[ybar,fill=blue,postaction={
		pattern=north east lines
	}] coordinates {%
		(Right Curves,0.0209534644365)%
		(Left Curves,0.011496670345)%
		(Crossing,0.0173330059701)%
		(Overtaking,0.0132451921235)%
		(Bridge,0.00407472034921)%
		(Steep Street,0.0256663702631)%
		(Bumpy Road,0.000316705165214)%
	};%
	%
	\addplot[ybar,fill=blue] coordinates {%
		(Right Curves,0.0266871218549)%
		(Left Curves,0.0422064081809)%
		(Crossing,0.0177071283795)%
		(Overtaking,0.105139685962)%
		(Bridge,0.0346626641884)%
		(Steep Street,0.00044493167196)%
		(Bumpy Road,0.0247823331152)%
	};%
	%
	\addplot[ybar,fill=cyan,postaction={
		pattern=north east lines
	}] coordinates {%
		(Right Curves,0.000211433045375)%
		(Left Curves,0.0241700348183)%
		(Crossing,0.0218363478874)%
		(Overtaking,0.377592278835)%
		(Bridge,0.0929346367002)%
		(Steep Street,0.0334709968863)%
		(Bumpy Road,0.00346031206002)%
	};%
	%
	\addplot[ybar,fill=cyan] coordinates {%
		(Right Curves,0.0805962357148)%
		(Left Curves,0.0466973984902)%
		(Crossing,0.0771217510504)%
		(Overtaking,0.112390691427)%
		(Bridge,0.00142221329715)%
		(Steep Street,0.033841344822)%
		(Bumpy Road,0.0821897618305)%
	};%
	%
	\addplot[ybar,fill=magenta,postaction={
		pattern=north east lines
	}] coordinates {%
		(Right Curves,0.414936899158)%
		(Left Curves,0.353582647599)%
		(Crossing,0.032929033973)%
		(Overtaking,0.0225146119977)%
		(Bridge,0.0482872028479)%
		(Steep Street,0.0264287373814)%
		(Bumpy Road,0.00115513324121)%
	};%
	%
	\addplot[ybar,fill=magenta] coordinates {%
		(Right Curves,0.125673548345)%
		(Left Curves,0.0167661286151)%
		(Crossing,0.0253009239147)%
		(Overtaking,0.250823107796)%
		(Bridge,0.0521581924772)%
		(Steep Street,0.0107519537798)%
		(Bumpy Road,0.0461338585328)%
	};%
	%
	
	\legend{
		Lancer (Baseline),Lancer (Ours),
		Lincoln (B.),Lincoln (O.),
		Smart (B.),Smart (O.),
		Golf (B.),Golf (O.),
		Van (B.),Van (O.)}
	
		\end{axis}%
		\end{tikzpicture}%
		
		\caption{Quantitative evaluation of the scale ratio of the methods \textit{constant distance} (ours) and \textit{intersection} (baseline). The provided values are the deviations w.r.t. reference scales.}
		\label{fig:evaluation:quantitative_scale_ratio}
\end{figure*}
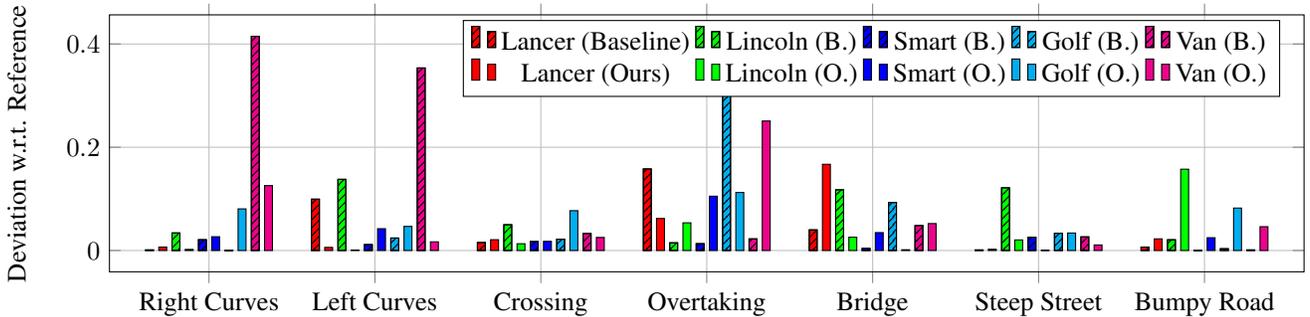

\begin{table*}
	\begin{center}
		\begin{tabular}{c|ccccc|ccccc}%
			\hline%
			Scale Ratio&\multicolumn{5}{c}{Average Scale Ratio Deviation w.r.t. Reference}&\multicolumn{5}{|c}{Average Trajectory Error}\\%
			Est. Type&Lancer&Lincoln&Smart&Golf&Van&Lancer&Lincoln&Smart&Golf&Van\\%
			\hline%
			Intersection (Baseline)&0.05&0.07&\textbf{0.01}&0.08&0.13&0.42&0.53&\textbf{0.25}&0.95&1.68\\%
			Constant Distance (Ours)&\textbf{0.04}&\textbf{0.04}&0.04&\textbf{0.06}&\textbf{0.08}&\textbf{0.20}&\textbf{0.23}&0.33&\textbf{0.33}&\textbf{0.47}\\%
		\end{tabular}%
	\end{center}
	\caption{Summary of the conducted evaluation. The second column shows the deviation of the estimated scale ratio w.r.t to the reference scale ratio per vehicle. The third column contains the average distances of the full dataset in meter. Overall, the trajectory error of the baseline and our approach is 0.77m and 0.31m.}
	\label{table:quantitative:evaluation_summary}
\end{table*}
\subsection{Scale Estimation Baseline: Intersection Constraints}
\label{subsection:scale_estimation_intersection}


The baseline is motivated by the fact, that some reconstructed points of the bottom of an object should lie in the proximity of the ground surface of the environment. Consider for example the reconstructed 3D points corresponding to the wheels of a car. This approach works only if at least one camera-point-ray of an arbitrary point in the object point cloud intersects the ground surface. For each camera we generate a set of vectors $\vec{v}_{j,i}^{(b)}$ pointing from the camera center $\vec{c}_{i}^{(b)}$ towards each object point $\vec{o}_{j,i}^{(b)}$. For non-orthogonal direction vectors $\vec{v}_{j,i}^{(b)}$ and normal vectors $\vec{n}_i$ we compute the ray plane intersection parameter for each camera-object-point-pair according to equation \eqref{eq:ray_plane_intersection_parameter}
\begin{equation}
\label{eq:ray_plane_intersection_parameter}
r_{j,i} = \frac{(\vec{p}_i -  \vec{c}_i^{(b)}) \cdot \vec{n}_i} { \vec{v}_{j,i}^{(b)} \cdot \vec{n}_i}.
\end{equation}
We compute the smallest ray-plane-intersection parameter for each image $i$.
\begin{equation}
\label{eq:r_per_image}
r_{i} = \min (\{r_{j,i} | j \in \{1, \dots, |\mathcal{P}^{(o)}|\}\})
\end{equation}
The intersection parameter $r_{i}$ corresponds to the point being closest to the ground surface, i.e. a point at the bottom of the object. Plugging $r_{i}$ in equation \eqref{eq:object_to_background_coordinates} for camera $i$ places the object point cloud on top of the ground surface. Thus, the smallest ray-plane-intersection-parameter $r_{i}$ is a value close to the real object-to-background-scale. Finally, we use the median $s$ of all image scale ratio factors as scale ratio factor to reconstruct the trajectory as computed in equation \eqref{eq:r_median}
\begin{equation}
\label{eq:r_median} 
r = med (\{r_{i} | i \in \{1, \dots, n_I \}\}),
\end{equation}
where $med$ denotes the median and $n_I$ the number of images.
We do not consider invalid image scale ratios $r_{i}$, i.e. cameras which have no camera-object-point-rays intersecting the ground representation. 

\subsection{Registration of Background Reconstruction and Virtual Environment}
\label{subsection:registration}

A common approach to register different coordinate systems is to exploit 3D-3D correspondences. To determine points in the virtual environment corresponding to background reconstruction points one could create a set of rays from each camera center to all visible reconstructed background points. The corresponding environment points are defined by the intersection of these rays with the mesh of the virtual environment. Due to the complexity of our environment model this computation is in terms of memory and computational effort quite expensive. Instead, we use the algorithm presented in \cite{Umeyama91} to estimate a similarity transformation $\vec{T}_s$ between the cameras contained in the background reconstruction and the virtual cameras used to render the corresponding video sequence. This allows us to perform 3D-3D-registrations of background reconstructions and the virtual environment as well as to quantitatively evaluate the quality of the reconstructed object motion trajectory. We use the camera centers as input for \cite{Umeyama91} to compute an initial reconstruction-to-virtual-environment transformation. Depending on the shape of the camera trajectory there may be multiple valid similarity transformations using camera center positions. In order to find the semantically correct solution we enhance the original point set with camera pose information, i.e. we add points reflecting up vectors $\vec{u}_i^{(b)} = \vec{R}^{T(b)}_i \cdot (0,1,0)^T$ and forward vectors $\vec{f}_i^{(b)} = \vec{R}^{T(b)}_i \cdot (0,0,1)^T$. For the reconstructed cameras, we adjust the magnitude of these vectors using the scale computed during the initial similarity transformation. We add the corresponding end points of up $\vec{c}_i^{(b)} + m \cdot \vec{u}_i^{(b)}$ as well as viewing vectors $\vec{c}_i^{(b)} + m \cdot \vec{f}_i^{(b)}$ to the camera center point set. Here, $m$ denotes the corresponding magnitude. 

\subsection{Reference Scale Ratio Computation}
\label{subsection:ground_truth_scale_ratio_computation}

As explained in section \ref{subsection:experiments_virtual_object_trajectory_evaluation} the presented average trajectory errors in Fig.~\ref{fig:evaluation:quantitative_trajectory_error} are subject to four different error sources. To evaluate the quality of the scale ratio estimation between object and background reconstruction we provide corresponding reference scale ratios. The scale ratios between object reconstruction, background reconstruction and virtual environment are linked via the relation shown in equation \eqref{eq:scale_ratios_ov}
\begin{equation}
\label{eq:scale_ratios_ov}
r_{(ov)} = r_{(ob)} \cdot r_{(bv)},
\end{equation}
where $r_{(ov)}$ and $r_{(bv)}$ are the scale ratios between object and background reconstructions and virtual environment, respectively.
The scale ratios $r_{(ob)}$ in Fig.~\ref{fig:evaluation:quantitative_scale_ratio} express the spatial relation of object and background reconstructions. 
The similarity transformation $\vec{T}_s$ defined in section \ref{subsection:registration} implicitly contains information about the scale ratio $r_{(bv)}$ between background reconstruction and virtual environment. To compute $r_{(ov)}$ we use corresponding pairs of object reconstruction and virtual cameras. We use the extrinsic parameters of the object reconstruction camera to transform all 3D points in the object reconstruction into camera coordinates. Similarly, the object mesh with the pose of the corresponding frame is transformed into the camera coordinates leveraging the extrinsic camera parameters of the corresponding virtual camera. The ground truth pose and shape of the object mesh is part of the dataset. In camera coordinates we generate rays from the camera center (i.e. the origin) to each 3D point $\vec{o}_{j}^{(i)}$ in the object reconstruction. We determine the shortest intersection $\vec{m}_{j}^{(i)}$ of each ray with the object mesh in camera coordinates. This allows us to compute $r_{(ov)}$ according to equation \eqref{eq:scale_ratios_ov_gt}
\begin{equation}
\begin{split}
\label{eq:scale_ratios_ov_gt}
r_{(ov)} =  med(  \{ 
med( \{ 
\frac{\lVert \vec{m}_{j}^{(i)} \rVert}{\lVert \vec{o}_{j}^{(i)} \rVert} 
|j \in \{1, \dots, n_J\} \}  \\
|i \in  \{1, \dots, n_I\} \} ).
\end{split}
\end{equation}
This allows us to compute the reference scale ratios $r_{(ob)}^{ref}$ using equation \eqref{eq:scale_ratios_ob_ref}
\begin{equation}
\label{eq:scale_ratios_ob_ref}
r_{(ob)}^{ref} = r_{(ov)} \cdot {r_{(bv)}}^{-1}. 
\end{equation}
Equation \eqref{eq:scale_ratios_ob_ref} shows that $r_{(ob)}^{ref}$ depends on the quality of the cameras in the background reconstruction and may slightly differ from the true scale ratio.

\section{Conclusions}	
\label{section:conclusions}
 
This paper presents a pipeline to reconstruct the three-dimensional trajectory of moving objects using monocular video data. We propose a novel constraint to estimate consistent object motion trajectories. In contrast to previous approaches, the presented scale estimation constraint is robust to occlusion and extents naturally to stationary objects. Due to the lack of 3D object motion trajectory benchmark datasets with suitable ground truth data, we present a new virtual dataset to quantitatively evaluate object motion trajectories. The dataset contains rendered videos of urban environments and accurate ground truth data including semantic segmentations, object meshes as well as object and camera poses for each frame. The proposed algorithm achieves an average reconstruction-to-ground-truth distance of 0.31 m evaluating 35 trajectories. In future work, we will analyze breakdown points of the proposed pipeline in more detail. This includes minimal object sizes, object occlusions and degeneracy cases. In addition, we intend to integrate previously published scale estimation approaches. These will serve together with the dataset\textsuperscript{\ref{project_page}} as benchmark references for future object motion trajectory reconstruction algorithms. 



{\small
	\bibliographystyle{ieee}
	\bibliography{egbib}
}

\end{document}